\documentclass{article} % For LaTeX2e
\usepackage{iclr2026_conference,times}

\usepackage{amsmath,amsfonts,bm}

\def\eqref#1{equation~\ref{#1}}
\def\1{\bm{1}}

\DeclareMathAlphabet{\mathsfit}{\encodingdefault}{\sfdefault}{m}{sl}
\SetMathAlphabet{\mathsfit}{bold}{\encodingdefault}{\sfdefault}{bx}{n}

\usepackage{hyperref}
\usepackage{url}
\usepackage{array}
\usepackage{newfloat}
\usepackage{listings}
\usepackage{wrapfig}
\usepackage{rotating}
\usepackage{subfigure}

\usepackage{wrapfig}
\usepackage{rotating}
\usepackage{subfigure}
\usepackage{booktabs}
\usepackage{enumerate}
\usepackage{multirow}
\usepackage[shortlabels]{enumitem}
\usepackage{color,soul}

\newcommand{\scheduler}{\textsc{Mixture-Scheduler}}
\newcommand{\sysname}{\textsc{TopoGen}}
\newcommand{\G}{$G$}
\newcommand{\bc}{$\mathbf{c}$}
\newcommand{\bz}{$\mathbf{Z}$}

\newcommand{\A}{$\mathbf{A}$}
\newcommand{\bzg}{$\mathbf{Z_G}$}
\newcommand{\bzc}{$\mathbf{Z_c}$}
\newcommand{\prior}{$p_{\theta}$}
\newcommand{\likelihood}{$p_{\psi}$}
\newcommand{\posterior}{$q_{\phi}$}
\newcommand{\evalmetric}{SD}

\usepackage{xcolor}

\title{Fine-Grained Graph Generation through Latent Mixture Scheduling}
\author{Nidhi Vakil \\
  Department of Computer Science \\
  University of Massachusetts Lowell \\
  \texttt{nidhipiyush\_vakil@uml.edu} \\ \And
  Hadi Amiri \\
  Department of Computer Science \\
  University of Massachusetts Lowell \\
  \texttt{hadi\_amiri@uml.edu} \\}
\iclrfinalcopy % Uncomment for camera-ready version, but NOT for submission.
\begin{document}

\maketitle

\begin{abstract}
Structure aware graph generation aims to generate graphs that satisfy given topological properties. It has applications in domains such as drug discovery, social network modeling, and knowledge graph construction. Unlike existing methods that only provide coarse control over graph properties, we introduce a novel conditional variational autoencoder for fine-grained structural control in graph generation. The approach refines the decoder’s latent space by dynamically aligning graph- and property-driven representations to improve both graph fidelity and control satisfaction. Specifically, the approach implements a mixture scheduler that progressively integrates graph and control priors. Experiments on five real-world datasets show the efficacy of the proposed model compared to recent baselines, achieving high generation quality while maintaining high controllability.
% Controlled graph generation aims to generate graphs that satisfy given topological properties. It has applications in domains such as drug discovery, social network modeling, and knowledge graph construction.
% Existing methods are limited in terms of fine-grained control, typically limiting generation to basic graph properties.
% This paper introduces \sysname{}, a novel conditional variational autoencoder for precise structural control in graph generation. \sysname{} refines the decoder's latent space by dynamically aligning structural and property-driven representations to improve both graph fidelity and control satisfaction. To further improve generation accuracy, we propose \scheduler, a new scheduling mechanism that progressively integrates graph and control priors.
% Experiments on five real-world datasets show the efficacy of \sysname{} compared to recent baselines, achieving lower graph edit distance and spectral errors while maintaining high controllability. We show that the improvements are due to our novel use of adjacency information during training and effective integration of graph and control representations. 
\end{abstract}
\section{Introduction}

Graph generation is a fundamental task in machine learning for modeling real-world networks such as molecular structures and social networks. Traditional models focus on producing graphs that follow general structural patterns (e.g. power-law distribution in node degree)~\citep{Erdos:1959:pmd,barabasi1999emergence,pmlr-v80-you18a}. However, many real world applications require {\em controlled} graph generation, where generated graphs must satisfy specific topological properties or attributes~\citep{zahirnia2024neural,martinkus2022spectre}. This is particularly crucial in domains such as 
drug discovery (generating new molecules that satisfy certain chemical properties)~\citep{jin2018junction,shi2019graphaf,jin2020hierarchical,luo2021graphdf,popova2019molecularrnn,Shi2020GraphAF,liu2021graphebm,zang2020moflow,de2018molgan}, 
synthetic material design~\citep{wang2022deep,sanchez2018inverse}, 
social and information networks~\citep{pitas2016graph,zhou2020data,zeno2021dymond}, 
knowledge graphs~\citep{melnyk-etal-2022-knowledge,DBLP:conf/emnlp/ZhouZG0023,DBLP:conf/emnlp/CaoH0LXLJZ23}, and 
programming languages (generating program graphs from source codes)~\citep{allamanis2018learning}.\looseness-1
% using abstract syntax trees and 

% Effective neural graph generation methods have been proposed. In particular, \citet{pmlr-v80-you18a}~and~\citet{li2018learning} proposed to turn input graphs into sequences and generate graphs sequentially using two recurrent neural networks, which iteratively determine if and how new nodes and edges should be added to the partially generated graph. Several effective methods have been developed specifically for text graph generation~\citep{koncel-kedziorski-etal-2019-text,
% jin-etal-2020-genwiki,guo2020interpretable,yao2020heterogeneous,song-etal-2020-structural,
% % wang-etal-2021-buildin,
% saha-etal-2022-explanation,melnyk-etal-2022-knowledge,han-shareghi-2022-self,edwards2022translation}. For example, \citet{guo2020interpretable} proposed a variational autoencoder that generates graphs through disentanglement of node and edge features. The model implements three encoders to model the distribution of node, edge and graph features; and two decoders to jointly generate these features based on the resulting latent representations.
% \citet{han-shareghi-2022-self} proposed a low-resource approach to generate text from input graph by introducing several graph masking strategies to integrate both local and global information into the pre-trained language models. 

Despite significant progress in graph generation, existing controlled graph generation methods are limited in scope, often restricting control to basic graph attributes 
such as node and edge counts 
as opposed to more fine-grained structural constraints, and lack a principled way to balance the 
structural 
generation process and attribute-based constraints.
For example, EDGE~\cite{chen2023efficient} is a discrete diffusion model that explicitly focuses on node degrees to control graph generation;
DiGress~\cite{vignacdigress} also builds on discrete diffusion techniques to 
% mainly focus to 
incorporate %specific graph 
% edge deletion invariant topological 
properties such as planarity or acyclicity for generating graphs;
Spectre~\cite{martinkus2022spectre} is a generative adversarial network that control graph generation by focusing on eigenvalues and eigenvectors, which provide abstract control over topological properties; and 
GenStat~\cite{zahirnia2024neural} is a variational autoencoder that learns a latent adjacency matrix from attributes such as number of edges, triangles,
%random walks histogram of specific length,
and $k$-hop neighbors histogram. Other works such as ~\citep{yang2019conditional,ommi2022ccgg} uses class labels and other class information as a condition to generate graphs where as ~\citep{liu2024graph,mercatali2024diffusion} focuses on molecule generation tasks.

% Existing methods either rely on latent representations which are difficult to interpret and control, focus on only a subset of graph attributes, or lack a principled way to balance structural and attribute-based constraints.

We propose \sysname{}, a novel conditional variational autoencoder for controlled graph generation based on fine-grained topological attributes. 
\sysname{} uses both the adjacency matrix and desired topological attributes during training for better latent space alignment and improved decoder tuning, {\em while relying only on attributes during inference}. We propose a novel scheduling mechanism (\scheduler) that progressively integrates structure- and attribute-driven latent representations for adaptive and precise control over generated graphs. \sysname{} provides flexibility in graph generation by generalizing to any number of fine-grained control attributes.

% \sysname{} implements an effective scheduling technique that integrates representations from both adjacency matrix and attribute distributions to enable more fine-grained and precise control for graph generation.\looseness-1 

% \sysname{} uses to combine graph representations with attribute representations to obtain infused representation  for the graph generator to control desired graph. 

% The closest approach to ours is GenStat~\citep{zahirnia2024neural}, which is a standard autoencoder model for controlled graph generation. It encodes given graph attributes to produce a latent adjacency matrix, which is then used by a decoder to produce the attributes. \sysname{} differs from GenStat from several aspects: it uses both graph attributes and the adjacency matrix of graphs {\em during} training for improved decoder tuning, while relying only on attributes during inference;  it introduces a novel scheduling technique to integrate latent representations from adjacency matrix and attribute distributions for effective training; and, unlike previous approaches, it can handle any number of fine-grained control attributes without any modification, to provide flexibility in graph generation.\looseness-1
% (ii) it uses a novel training paradigm of using adjacency matrix from the graph as an additional information during training to better tune the decoder, and
% theoretically-driven
The contributions of this work are:
\begin{itemize}
    \item \sysname{}: a novel conditional variational autoencoder that enables fine-grained topological control using both graph adjacency matrices and attribute vectors during training, while relying only on attributes during inference for precise graph generation.\looseness-1
    \item \scheduler{}: a latent space integration technique to dynamically balance adjacency matrix and attribute representations for generation.\looseness-1
\end{itemize}
We compare \sysname{} against current models for controlled graph generation on several datasets and generation tasks. 
Our key findings are as follows:
(1) Joint adjacency-attribute integration improves generation quality, aligning graphs with specific attribute constraints more effectively than prior models;
(2) Gradual incorporation of prior information (from attribute via \scheduler) during training improves controlled graph generation; 
(3) Increasing the number of control attributes improves generation precision, which confirms that fine-grained constraints help generate structurally valid graphs. Our code and data will be released.\looseness-1

\section{Controlled Graph Generation}\label{sec:cgg}

\paragraph{Problem Definition}
Given a vector $\mathbf{c}$ that represents the fine-grained topological attributes of a target graph \G{}, our goal is to generate a graph $\hat{G}$ whose structure satisfies the attributes $\mathbf{c}$.

\begin{figure*}
    \centering
    \includegraphics[scale = 0.35]{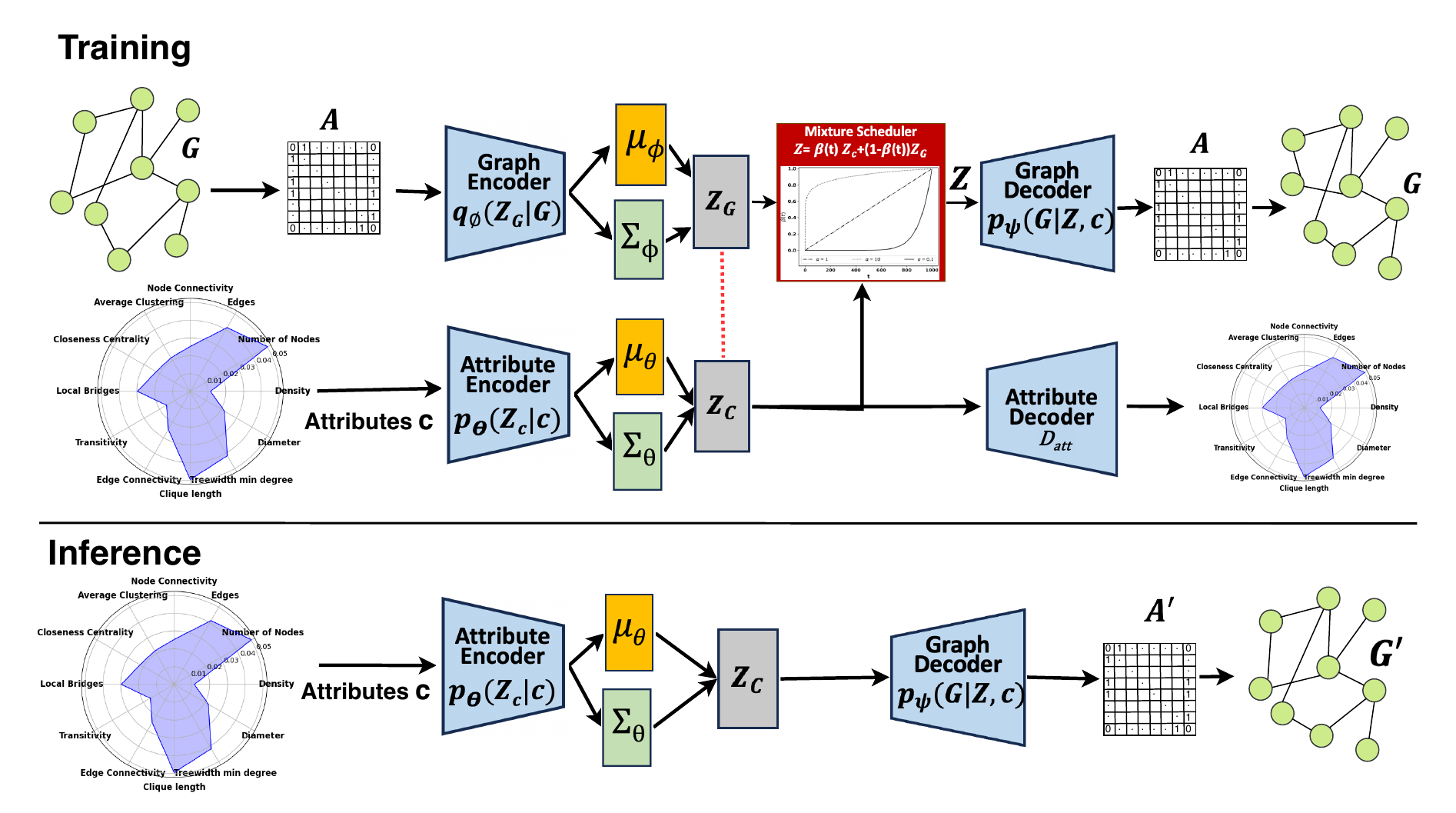}
    \vspace{-10pt}
    \caption{\sysname{} uses both graph attributes and adjacency matrix {\em during} training for improved decoder tuning. It implements a novel scheduling technique to effectively integrate attributes and graph distributions to provide fine-grained topological control in generation. At test and inference times, it only relies on desired attributes to generate graphs.}
    \label{fig:arch}
\end{figure*}

\paragraph{Solution Overview} 
We formulated the above problem as a ``learning to generate'' task. During {\em training}, \sysname{} uses the adjacency matrix \A{} of the target graph \G = $(V,E)$ and its corresponding attribute vector $\mathbf{c}$ to learn the joint distribution of graphs and their attribute vectors for controlled graph generation. 
As Figure~\ref{fig:arch} shows, \sysname{} encodes the structural representation \bzg{} from adjacency matrix \A{} to parameterize the posterior distribution $q_\phi$, and 
the attribute representation \bzc{} from the attribute vector $\mathbf{c}$ to define the prior distribution $p_\theta$. 
% \sysname{} samples from distributions $q_\phi$ and $p_\theta$ to combines \bzg{} and \bzc{} 
These representations are combined using \scheduler{} to balance structural and attribute information in the latent representation \bz{}. The \scheduler{} aims to align $q_{\phi}$ and $p_\theta$, as they both represent graphs with the same topological structure. 
The decoder then learns the likelihood distribution $p_{\psi}$ from \bz{} to generate a graph $\hat{G}$ that satisfies the specified attributes $\mathbf{c}$. 
% \sysname{} uses both graph and attribute during training to tune the parameters for all the distributions. 
At {\em inference} time, \sysname{} generates graphs using only the prior $p_{\theta}$ and the likelihood $p_{\psi}$, conditioned solely on the attribute vector $\mathbf{c}$.
% At the inference time, however, \sysname{} only uses the control attribute vector to generate graphs that satisfy the given attribute vectors. 
% To simplify the generation process, we assume a maximum threshold for the number of nodes in generated graphs.
% \paragraph{Findings:}

\paragraph{Control Attributes} 
We provide a list of structural attributes for explicit and precise control over the graph generation process. These include 
\textbf{number of nodes \& edges}, which define the scale of the target graph; 
\textbf{number of local bridges}, which is the number of edges that are not part of a triangle in the graph, these ``bridges'' transfer information between different graph regions;
\textbf{graph density}, which is the fraction of possible edges in the graph, computed as $e/v(v-1)$, where $e$ is the number of edges and $v$ is the number of nodes in the graph;
\textbf{edge connectivity}, which is the minimum number of edges that must be removed to disconnect the graph; 
\textbf{node connectivity}, which is the minimum number of nodes that must be removed to disconnect the graph;
\textbf{number of maximum cliques}, which is the number of maximal complete subgraphs in the graph;
\textbf{graph diameter}, which is the length of the shortest path between the most distanced nodes in a graph;
\textbf{treewidth min degree}, which is an integer quantifying how much the graph deviates from a tree;
\textbf{closeness centrality}, which is the average distance of a node to all other nodes in its corresponding connected component, averaged across all nodes;
\textbf{clustering coefficient}, which is the fraction of triangles within a node's immediate neighbors, averaged across all nodes; and 
\textbf{transitivity}, which is the fraction of all possible triangles present in a graph, computed as $3\times |triangles|/|triads|$, where a ``triad'' is a set of three nodes connected by at least two edges.
\paragraph{Importance:}These attributes enable precise control over graph generation and make it possible to generate graphs that satisfy diverse and complex structural requirements. Attributes such as transitivity and graph density can be adjusted to manage the connectivity of graphs. For example, increasing graph density can improve the robustness of local area networks in terms of reliable communication, fine-tuning transitivity can help model molecular structures with specific bonding properties, or adjusting transitivity can help simulate disease spread patterns in human contact networks in epidemiology. Granular control over these attributes allows for generating graphs that satisfy specific needs across various applications. This includes creating balanced graph datasets with controlled structural diversity;
augmenting small-scale datasets by generating similar yet distinct subgraphs, especially in fields like medicine, where obtaining large-scale real-world data is expensive or infeasible; and finding novel structures in fields such as chemistry and molecular biology. For example, in drug discovery, the generation of graphs that represent potential compounds with desired properties can accelerate the search for and discovery of new therapeutics.

\subsection{Graph Encoding through Mixture Scheduling}
We introduce a new approach to graph encoding by gradually balancing structural (i.e. adjacency matrix) and attribute-based representations during training. Unlike conventional methods that rely on direct sampling or divergence minimization (e.g., Wasserstein distance or KL divergence), our approach dynamically controls the contribution of structural and attribute representations using a smooth scheduling function. This allows for flexible and adaptive representation learning, where generated graphs preserve topological properties as well as align with desired attribute constraints. 
\paragraph{Graph Encoder}
\sysname{}'s encoder uses a convolution neural network (CNN) to encode the structural information of graph \G{} into a latent representation \bzg{} using $s$ channels, and parameterize the posterior distribution \posterior. This distribution is defined as:
\begin{eqnarray}
q_{\phi}\left(\mathbf{Z_G}|G\right) & = & \mathcal{N}\left(\mathbf{Z_G}|\mu_{\phi}=h\left(G\right),\Sigma_{\phi}=h'\left(G\right)\right),
\end{eqnarray}
% neural network
where $\mathcal{N}$ is a Gaussian distribution with mean vector $\mu_{\phi}=h\left(G\right)$ and covariance matrix $\Sigma_{\phi}=h'\left(G\right)$, both obtained from the CNN  with parameters $\phi$, where 
$h(G)$ computes the mean vector using features from the first half ($s/2$) of CNN channels and $h'(G)$ computes covariance matrix from the second half ($s/2$) of the CNN channels.  
The partitioning is similar to how variational autoencoders (VAEs) separate their latent space into a mean and variance to generate diverse samples while preserving meaningful structure. It allows different parts of the CNN to capture distinct statistical properties of the latent space, explicitly control uncertainty and variability, and encode well-structured representations.
While \sysname{} is compatible with GNNs, we focus on CNNs due to better performance in our experiments. Other frameworks have used Multilayer Perceptron (MLP) layers~\cite{zahirnia2024neural,vignacdigress,jograph} or a combination of LSTM, MLP and message passing~\cite{chen2023efficient} as graph encoder.

\paragraph{Attribute Encoder}
To control graph generation, the attribute encoder in Figure~\ref{fig:arch} learns the representation of the attribute  $\mathbf{c}$,
% , such that similar attribute vectors are also similar in the latent space. 
and the parameters for the prior distribution \prior{} are learned as follows:\looseness-1
% using a normal distribution with the mean obtained from a non-linear transformation of $\mathbf{c}$:
\begin{eqnarray}
p_{\theta}\left(\mathbf{Z_c}|\mathbf{c}\right) & = & \mathcal{N}\left(\mathbf{Z_c}|\mu_{\theta} = f\left(\mathbf{c}\right),\Sigma_{\theta} = I\right),
\end{eqnarray}
where $f(\mathbf{c})$ is a non-linear transformation of the attribute vector from a feed forward neural network to capture interaction between the features of the attribute representation, and $\Sigma_{\theta}$ is the unit variance.

\paragraph{Mixture Scheduler}
Unlike conventional approaches that align prior and posterior distributions using Wasserstein distance~\citep{d2eb0123} or divergence techniques~\citep{kullback1951information}, we introduce \scheduler{}, a principled approach that gradually integrates the prior \prior{} and posterior \posterior{} to learn effective representations that satisfy desired attribute \bc. Instead of abrupt transitions, \scheduler{} enables a smooth and adaptive interpolation between structural and attribute-based latent representations. We define the final latent representation as:
\begin{equation}\label{eq:mixture}
    \mathbf{Z} = \beta(t) \mathbf{Z_c} + \left(1 - \beta(t) \right) \mathbf{Z_G},  
\end{equation}
where $\beta(t)$ is the {\em inclusion factor} at epoch $t$, which controls the gradual incorporation of the prior $\mathbf{Z_c}$ during training. To derive a general form of $\beta(t)$, we assume that the rate by which the prior $\mathbf{Z_c}$ is incorporated is uniformly distributed over the remaining training time:\looseness-1
\begin{equation}
    \frac{d\beta(t)}{dt} = \frac{1-\beta(t)}{1-t},
\end{equation}
where $t\in[0,1]$ is normalized training progress, with $t=1$ when $\mathbf{Z_c}$ is fully incorporated. Solving this differential Equation, we obtain:
\begin{equation}
    \small
    \int \frac{1}{1-\beta(t)}d\beta(t) = \int \frac{1}{1-t}d(t),
\end{equation}
which results in $\beta(t) = 1-\exp(c)(1-t)$ for some constant $c$. Setting the initial inclusion value as $\beta(0)$ (at $t=0$) and $\beta(1) =1$, we obtain a {\em linear} scheduler:
\begin{equation}
    \beta(t) = \min\big(1, 1-(1-\beta(0))(1-t)\big).
\end{equation}

We modify the linear scheduler to allow for adaptive control over the rate at which $\mathbf{Z_c}$ is incorporated at different training stages. This results in the {\em generalized inclusion function}:
\begin{equation}\label{eq:scheduler}
    \beta(t) = \min\left(\gamma, \left(1-(1-\beta(0)) (1- t)\right)^{\frac{1}{\alpha}}\right),
\end{equation}
\begin{wrapfigure}{r}{0.4\textwidth}
    \vspace{-15pt}
    \centering
    \includegraphics[scale=0.45]{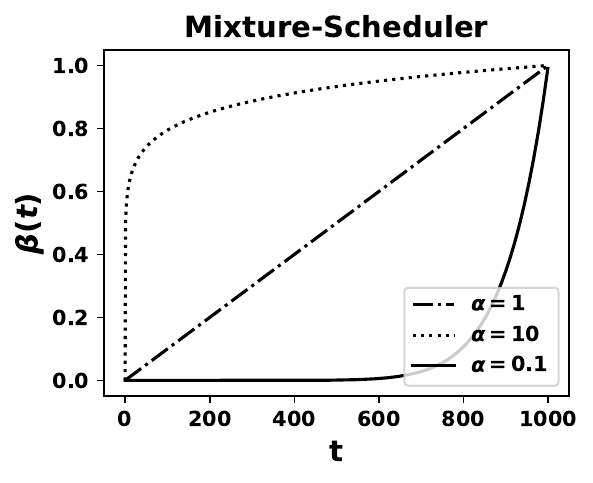}
    \vspace{-15pt}    
    \caption{The parameter $\alpha$ controls the inclusion factor, ($\beta(t)$) in (\ref{eq:mixture}). It specifies how quickly the prior is integrated during training. A smaller $\alpha$ results in less inclusion of \prior{} during the early training epochs, with gradual inclusion increasing toward the end of training.\looseness-1}
    \label{fig:mixture_scheduler}
    \vspace{-20pt}    
\end{wrapfigure}
where $\gamma\in[0,1]$ controls the maximum possible inclusion from prior \prior{}; 
$\alpha>0$ determines the rate at which the prior is integrated during training, see Figure~\ref{fig:mixture_scheduler}; $t$ represents the current epoch; and $\beta(0)$ is the initial inclusion value. \looseness-1
% \begin{figure}
The intuition behind developing (\ref{eq:scheduler}) is to provide flexible control over the contributions of the prior and posterior and allow for smooth and gradual transition between them; see Figure~\ref{fig:mixture_scheduler}. By gradually increasing the influence of $\mathbf{Z_c}$, the learned representations retain meaningful graph topology while aligning with the desired attribute constraints.

% We found that using mixture from both the distribution helps the estimator to learn representation such that it brings the graphs with similar topological structure closer and also align with corresponding attribute representations (\ref{sec:scheduler_ablation}).

% \paragraph{Theoretical Justification} 
% \hadi{TODO} 
The \scheduler{} can be understood as a \textit{soft optimization constraint} that enables smooth interpolation between probability distributions. It dynamically transitions the latent representation $\mathbf{Z}$ from the structural posterior $q_{\phi}(\mathbf{Z_G} | G)$ to the attribute-conditioned prior $p_\theta(\mathbf{Z_c} | \mathbf{c})$ during training. 
% first learning structural properties and then gradually enforcing the attribute constraints.
The scheduler implicitly minimizes the Wasserstein distance ($W$) between the structural and attribute-driven distributions
% \begin{equation}
%     W\left(q_{\phi}(\mathbf{Z_G} | G), p_\theta(\mathbf{Z_c} | \mathbf{c})\right) \leq f(\alpha, \gamma, t),
% \end{equation}
% where $f(\alpha, \gamma, t)$ defines an upper bound on the divergence, 
controlled by the scheduling parameters $\alpha$ and $\gamma$, to govern the transition dynamics:
a smaller $\alpha$ results in a slow transition, i.e. prioritizing structural learning before enforcing attribute constraints; 
a larger $\alpha$ causes a faster shift towards $\mathbf{Z_c}$, i.e. aligning graphs with attributes earlier but risking instability; and 
$\gamma$ controls the final alignment of $\mathbf{Z}$ with $\mathbf{Z_c}$. Higher $\gamma$ enforces stronger attribute constraints but may distort structural properties.
%
% We expect early reliance on \(\mathbf{Z_G}\) enables the model to capture structural patterns before being guided by attribute constraints. The gradual shift towards \(\mathbf{Z_c}\) aligns the generated graphs with specified attributes without losing structural consistency. The scheduler prevents abrupt latent space shifts to reduce divergence-related training instabilities.

\subsection{Attribute-Guided Graph Generation}
\sysname{} introduces a novel attribute-guided graph generation framework. Unlike conventional methods that rely solely on structural embeddings, our framework incorporates graph attributes as constraints to enable precise controlled graph generation. Using a Bernoulli-based likelihood model, we allow flexible edge prediction while maintaining topological consistency. In addition, we introduce a distance-regularized objective function to enforce smooth transitions between prior and posterior distributions to balance structural fidelity with attribute adherence. 

\paragraph{Graph Generation}
We model graph generation using a Bernoulli distribution \cite{Murphy2012} to determine edge probabilities between node pairs and generate the adjacency matrix \A{}. The graph decoder learns the likelihood  distribution $p_\psi$ from \bz{} to maximize the probability of generating graphs that satisfy the attribute constraints \bc{}: 
\begin{eqnarray}
p_\psi\left(G|\mathbf{Z},\mathbf{c}\right) \sim \text{Bernoulli}\left(D_{graph}\left(\mathbf{Z}\right)\right),
\end{eqnarray}
where $\mathbf{Z}$ represents the latent representation processed by the decoder $D_{graph}$ to obtain the parameters of the Bernoulli distribution. Here, a value of 1 from the Bernoulli distribution indicates an edge between a node pair.\looseness-1

\paragraph{Training Objective}
We develop the following objective function to learn model parameters:
\begin{align}
\label{eq:elbo}
\mathcal{L}&(\phi,\theta,\psi|G,\mathbf{c}) 
    = \\ \nonumber
    & \underbrace{\mathbb{E}_{q_{\phi}(\mathbf{Z}|G)}\left[\log p_{\psi}(G|\mathbf{Z},\mathbf{c})\right]}_{graph-reconstruction} - \lambda_{d} \cdot \underbrace{\mathcal{D} \left(q_{\phi}(\mathbf{Z_G}|G), p_\theta(\mathbf{Z_c}|\mathbf{c})\right)}_{distance-function}  - \lambda_{c} \cdot \underbrace{\mathbb{E}_{p_{\theta}(Z_{c}|\mathbf{c})} \left[\left(\mathbf{c} - D_{\text{att}}(Z_{c})\right)^{2}\right]}_{attribute-reconstruction},
\end{align}
% \begin{eqnarray}
% \label{eq:elbo}
% \mathcal{L}\left(\phi,\theta,\psi|G,\mathbf{c}\right) = \mathbb{E}_{q_{\phi}\left(\mathbf{Z}|G\right)}\left[\log p_{\psi}\left(G|\mathbf{Z},\mathbf{c}\right)\right]  \\\nonumber
% -\lambda_{distance}\cdot \mathcal{D}\left(q_{\phi}\left(\mathbf{Z_G}|G\right),p_\theta\left(\mathbf{Z_c}|\mathbf{c}\right)\right) \\\nonumber
% + \lambda_{c}\cdot \mathbb{E}_{p_{\theta}\left(Z_{c}|c\right)}\left[\left(c-D_{att}\left(Z_{c}\right)\right)^{2}\right],
% \end{eqnarray}
where the \textbf{first term} %expected log-likelihood term 
is the reconstruction loss, which encourages generating graphs that are structurally similar to the given graph \G{}, conditioned on the latent representation \bz{} and attributes \bc{}.
The \textbf{second term} ($\mathcal{D}$) is a general distance function for probability distributions; it regularizes the objective by computing the difference between the approximate posterior $q_{\phi}(\mathbf{z}|G)$ and the prior $p_\theta(\mathbf{z}|\mathbf{c})$ to explicitly enforce alignment between learned graph structures and attribute-driven representations. 
We note that \scheduler{} implicitly aligns the posterior (\(\mathbf{Z_G}\)) and prior (\(\mathbf{Z_c}\)).
% However, relying solely on the scheduler may lead to instability in latent space alignment, especially early in training when the model has not yet learned meaningful representations. 
Without explicit regularization, \(\mathbf{Z_G}\) and \(\mathbf{Z_c}\) may remain disjoint, and result in poor attribute-guided graph generation. The second term provides an explicit constraint for smoother transitions, prevents posterior drift, and stabilizes training by enforcing gradual alignment between structural and attribute-driven latent spaces.
The distance function, $\mathcal{D}$, can be chosen based on application needs. We used the Wasserstein distance due to its symmetric property. However, other distance functions can also be used. 
The \textbf{third term} encourages accurate reconstruction of the attribute vector $\mathbf{c}$, using a neural network based attribute decoder $D_{att}()$, described below.
$\lambda_{d}$~and~$\lambda_{c}$ are hyperparameters to balance these terms.

\paragraph{Attribute Decoder}
During training, we use a feedforward neural network as the attribute decoder to reconstruct attributes from latent representation \bzc{}. To guide accurate graph generation that aligns with the specified control attributes, we minimize the mean square error (MSE) between the ground truth and predicted attribute vectors and effectively guide the model toward attribute-consistent generation; see the third term in (\ref{eq:elbo}).

\paragraph{Inference Process}
During inference, the model generates a graph conditioned on the desired attribute vector \bc{} using the prior distribution \prior{}, as illustrated in Figure~\ref{fig:arch}. The prior is first used to create a latent representation, which encodes attribute-driven structural properties. This representation is then passed to the decoder to parameterize the \likelihood{} distribution to sample and generate a graph that satisfies the specified attributes. Unlike training, inference relies only on the prior, so that graph generation is fully controlled by the desired attributes without requiring reference graphs.

\section{Experiments}
\paragraph{Datasets} 
We use several datasets for experiments:
% \paragraph{\texttt{WordNet}}~\citep{miller1995wordnet}:
{\bf WordNet}~\citep{miller1995wordnet}: a large lexical dataset of English, where words are grouped into synonym groups (synsets) and are connected by linguistic 
\begin{wraptable}{r}{0.38\textwidth}
\vspace{-10pt}
\footnotesize
\centering
\caption{Dataset statistics in terms of number of graphs.}
\begin{tabular}{llll}
              \textbf{Dataset}  & \textbf{Train} & \textbf{Val}   & \textbf{Test}\\
         \toprule
        \textbf{WordNet} &    52,675  & 2,926   & 2,927   \\
        \textbf{Citeseer}&    1,406   &   78   & 79 \\
        \textbf{Arxiv}   &    47,538  & 2,641   & 2,641 \\
        \textbf{MUTAG} & 169 & 10 & 9\\
        \textbf{MOLBACE} &1,323 & 74 &74\\
        \bottomrule
\end{tabular}
\label{tab:datasets}
\vspace{-5pt}
\end{wraptable}
relationships. We construct four distinct WordNet graphs using hypernyms, hyponyms, meronyms, and holonyms relations. 
{\bf Ogbn-arxiv}~\citep{hu2020open}: The Open Graph Benchmark dataset includes a citation network of computer science papers from arXiv, with nodes as papers and edges represent citations among papers. Each paper carries an embedding derived from its title and abstract. 
{\bf Citeseer}~\citep{kipfW17}: a citation network of scientific articles, where nodes are papers and edges indicate citations between them.
{\bf MUTAG}~\citep{DBLP:journals/corr/abs-2007-08663}: a molecular dataset where each graph represents a chemical compound labeled based on its mutagenic effect on specific gram negative bacterium.
{\bf MOLBACE}~\citep{hu2020open}: a molecular dataset where each graph represents a chemical compound.  
We create several datasets of graphs by extracting $k$-hop neighbors, $k=\{2,3\}$, around each node in the above graphs to create training, validation and test data splits for controlled graph generation. Table\ref{tab:datasets} shows the statistics of these datasets. 
\paragraph{Evaluation Metrics}
We compute the difference between predicted and ground truth graphs to compare models in controlled graph generation using two metrics: 
{\bf Graph Edit Distance (GED$\downarrow$)}~\cite{sanfeliu1983distance}: is a structural similarity (or dissimilarity) measure that quantifies the minimum number of edit operations (node/edge insertions, deletions, or substitutions) required to transform one graph into another. It provides a fine-grained comparison by explicitly capturing structural differences. However, GED is computationally expensive, as finding the exact edit distance between two graphs is NP-hard~\cite{zeng2009comparing}. Therefore, it is often used to determine structural similarity among small graphs.
{\bf Spectral Difference (SD$\downarrow$)}~\citep{jograph}: a widely used approach for comparing structural properties of graphs. It uses the sorted eigenvalues of the Laplacian matrix, which encode global structural properties such as connectivity, clustering tendencies, and diffusion dynamics. Unlike node-to-node matching methods like GED, SD is invariant to node ordering, robust to small local perturbations, and computationally efficient. For fair and meaningful comparison between predicted and ground truth graphs of different sizes, we align their eigenvalue ($\lambda$) distributions by zero-padding the smaller graph's eigenvalues to match the size of the larger graph, and report average spectral difference, $SD=1/n \times ||\lambda_{groundtruth}-\lambda_{pred}||_2$ for each dataset. We chose SD and GED as evaluation metrics to specifically focus on fine-grained attribute fidelity and node-to-node alignment, which are most directly influenced by different models’ attribute-conditioning objectives. {\bf Maximum Mean Discrepancy (MMD$\downarrow$)}~\citep{pmlr-v80-you18a,vignacdigress} is widely used for distribution-level comparison; however, they do not provide fine-grained evaluation on individual attributes. MMD results are reported in Appendix. 
% \hadi{instead of this, which may not seem/be convincing, compute them at individual index level and report in the appendix, it's okay if the Table is big}
% We use Spectral gap metric (\evalmetric{}$\downarrow$) for evaluation which is the difference between eigenvalues of the graph Laplacian of predicted and ground truth graph. 
% MAD computes the absolute difference between the attributes of predicted graphs and their corresponding target graphs.
% GED is a graph similarity measure which provides the minimum cost for transforming one graph to another. Due to high time complexity, we only evaluate this metric for test graphs with 10 or less nodes.
%is similar to Levenshtein distance for strings. It
 % and graph edit distance (GED$\downarrow$)

\paragraph{Baselines} We compare \sysname{} against several recent baselines. For a fair comparison, we incorporate our control attributes into all models that support conditioning. The exception is GraphRNN, which is a free (non-controlled) generative model and provides a point of comparison for evaluating the benefits of attribute conditioning.
% \textit{Random}: Erd\H{o}s-R\'enyi graph with probability of edge creation set to the edge density of the gold graph. 
%
{\bf GraphRNN}~\citep{pmlr-v80-you18a}: generates graph iteratively by training on a representative set of graphs using breath first search of nodes and edges and implements node and edge RNNs to generate target graphs. GraphRNN is not a controlled generation approach.
{\bf EDGE}~\citep{chen2023efficient}: is a diffusion based generative model which iteratively removes edges to create a completely disconnected graph and uses decoder to iteratively reconstruct the original graph. 
It explicitly uses adjacency matrix to satisfy the statistics of the generated graphs during training. 
{\bf GenStat}~\citep{zahirnia2024neural}: learns the latent adjacency matrix conditioned on graph level attributes, and decodes it to recreate attribute statistics and use them to generate graphs.
{\bf DiGress} ~\citep{vignacdigress}: learns to generate graphs by discrete denoising diffusion model with categorical nodes and edge attributes and by incorporating graph-theoretic features.
{\bf GruM} ~\citep{jograph}: is a graph generation framework which captures the topology of the graph and predicts the graph using  mixture of endpoint-conditioned diffusion processes.
% is a graph generative framework that uses diffusion process to learn topology of the graph as a mixture of endpoint-conditioned  Ornstein-Uhlenbeck processes to drift the model in the direction of achieving topologically correct graph.

\begin{table}
\centering
\caption{Performance comparison across multiple datasets. We evaluate models using Spectral Difference (\evalmetric{}) (left) and Graph Edit Distance (GED) (right), where lower scores indicate better performance. All models are optimized using the same set of attribute constraints described in \S\ref{sec:cgg}.}
\resizebox{\textwidth}{!}{
\begin{tabular}{lccccc|ccccc} 
        & \textbf{WordNet}& \textbf{Citeseer}& \textbf{Ogbn-Arxiv} & \textbf{MUTAG} &\textbf{MOLBACE} & 
        \textbf{WordNet}& \textbf{Citeseer}& \textbf{Ogbn-Arxiv} & \textbf{MUTAG} &\textbf{MOLBACE} \\

        \toprule
         & \multicolumn{5}{c|}{\textbf{\evalmetric} ($\downarrow$)} & \multicolumn{5}{c}{\textbf{GED} ($\downarrow$)}\\
        \midrule

        % \textbf{Random}           & 17.12  &  16.20 & 19.53  & 11.42 & 15.44 & 34.24& 46.37&\underline{49.13} & 36.00& 66.3\\

        \textbf{GraphRNN}           & \textbf{0.31}  & 0.42  &  \underline{0.46} & \underline{0.21} &0.13  & 32.58& 54.83& 52.43& \underline{15.77}& \underline{41.52}\\
        \midrule

        \textbf{GenStat}  &  \underline{0.32} & \underline{0.40} & 0.47 & \underline{0.21} & \underline{0.11}  & 37.01& 58.69& 63.77&34.66 &61.32\\
        \textbf{EDGE}      &  0.32 & 0.44 & 0.47 & 0.27 & 0.14  & 35.16&59.58 & 59.70& 29.77&72.93\\
        \textbf{DiGress}       &  0.37 & 0.85 & 0.73 & 0.88 & 0.91 & 31.32&\textbf{45.16}& \textbf{47.24}& 32.66& 61.64\\
        \textbf{GruM}                & 0.40 & 0.43 & 0.50 & 0.52 & 0.69 & \underline{29.68}&55.06 & 51.87& 27.00& 65.27\\
        \midrule

        \textbf{\sysname{}}                     &  0.44 & \textbf{0.27} & \textbf{0.40} & \textbf{0.10} & \textbf{0.09} & \textbf{26.79}&\underline{45.62} &\underline{49.68} & \textbf{12.88}& \textbf{28.90}\\
         \bottomrule
\end{tabular}
}
\label{tab:result}
\vspace{-10pt}
\end{table}

\subsection{Main Results}
Table~\ref{tab:result} shows the overall performance of models across datasets. \sysname{} consistently achieves the lowest \evalmetric{} across all datasets and the lowest GED across 3 out of 5 datasets, which indicate more accurate structural alignment and better controlled graph generation. DiGress achieves the best GED on Citeseer and Ogbn-Arxiv, but struggles in SD and underperforms on domain specific datasets like MUTAG and MOLBACE.
GruM has competitive SD scores in WordNet, Citeseer, and Ogbn-Arxiv, but performs worse in GED.

The high GED performance of DiGress on citation networks (Citeseer and Ogbn-Arxiv) is perhaps because DiGress is optimized for handling dense and scale-free networks, whereas \sysname{} is designed for fine-grained attribute control and may not explicitly prioritize preserving connectivity hubs. In addition, DiGress likely preserves local citation patterns better than \sysname{}, leading to lower GED scores. 
In addition, GraphRNN achieves lower SD and GED scores on MUTAG and MOLBACE compared to most baselines. Unlike citation or social networks, molecular graphs have high local dependencies--atoms must be connected in precise ways to form valid molecules, where certain structures appear frequently (e.g., benzene rings, carbon chains). The sequential approach of GraphRNN perhaps better learns these recurring patterns, which makes it effective for generating realistic molecular graphs. Other baselines (DiGress and GruM) underperform on capturing the fine-grained rules that govern molecular connectivity.
Table~\ref{tab:visual_result} shows examples of different graphs generated by \sysname{} and GruM across datasets; see Appendix~\ref{app:vis}, Table~\ref{tab:other_visual_result} for outputs of other models.  As evident from the Table, \sysname{} generates graph that are more similar to the target graphs compared to other baseline models. We attribute this improvement to \sysname{}'s ability to perform fine-grained controlled generation using graph attributes. Generation error for each attribute is detailed in Appendix~\ref{sec:att_wise_analysis}. \looseness-1
\begin{table*}[t!]
    % \small
    \tiny
    \centering
    \caption{Graph visualization across datasets. Examples are taken from test splits of datasets. }
    \begin{tabular}{p{0.2cm} *{10}{>{\centering\arraybackslash}p{0.9cm}}} 
    % \begin{tabular}{p{0.2cm}*{10}{p{0.9cm}}}
         & \multicolumn{2}{c}{\textbf{Wordnet}} & \multicolumn{2}{c}{\textbf{Citeseer}} & \multicolumn{2}{c}{\textbf{Ogbn-Arxiv}} & \multicolumn{2}{c}{\textbf{Mutag}} & \multicolumn{2}{c}{\textbf{Molbace}}\\
         \toprule
         
         {\begin{turn}{90} Test \end{turn}} 
         & {\includegraphics[scale=0.2]{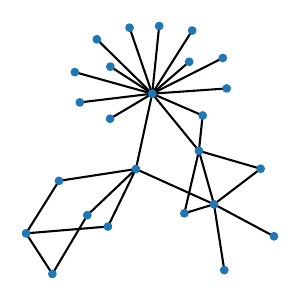}}  
         & {\includegraphics[scale=0.2]{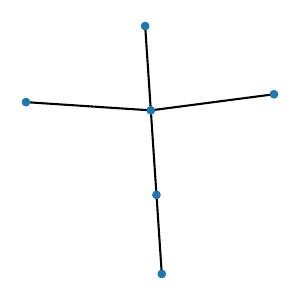}}
         & {\includegraphics[scale=0.2]{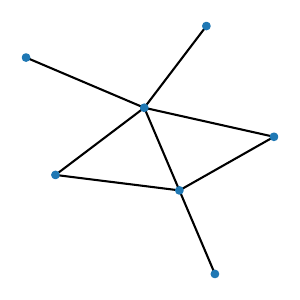}}
         & {\includegraphics[scale=0.2]{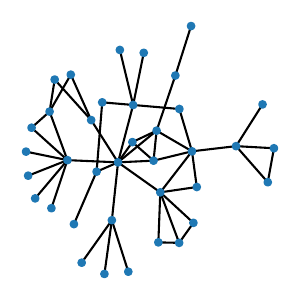}}
         & {\includegraphics[scale=0.2]{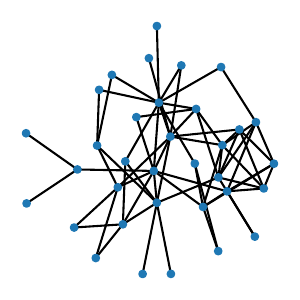}}
         & {\includegraphics[scale=0.2]{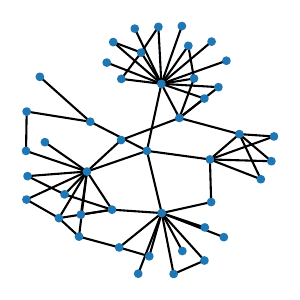}}
         & {\includegraphics[scale=0.2]{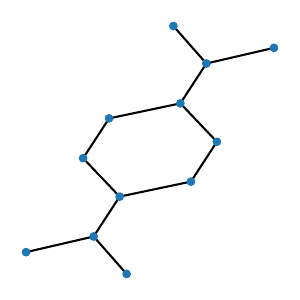}}
         & {\includegraphics[scale=0.2]{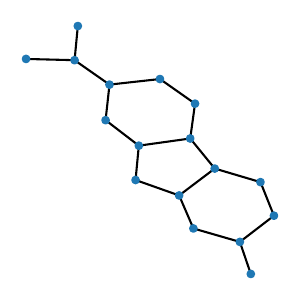}}
         &{\includegraphics[scale=0.2]{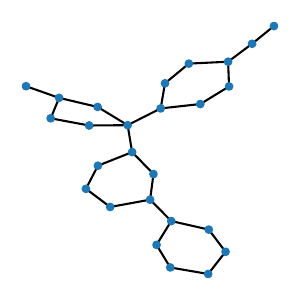}}
         & {\includegraphics[scale=0.2]{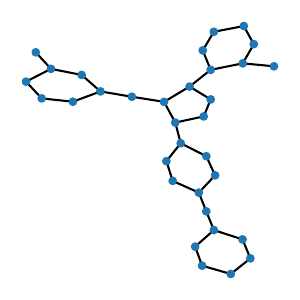}} \\
         \midrule
         
         {\begin{turn}{90}\sysname{} \end{turn} }
         & {\includegraphics[scale=0.2]{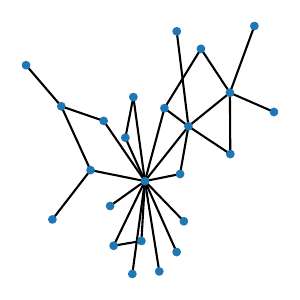}}  
         & {\includegraphics[scale=0.2]{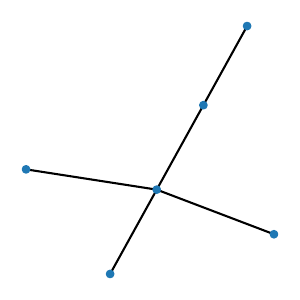}}
         & {\includegraphics[scale=0.2]{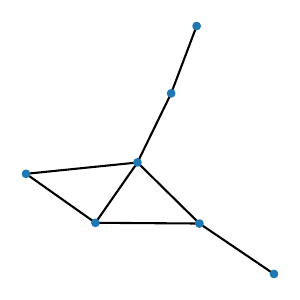}}
         & {\includegraphics[scale=0.2]{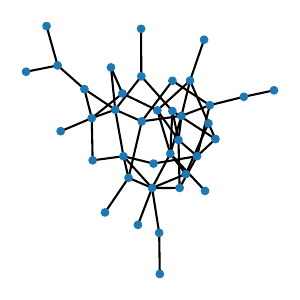}}
         & {\includegraphics[scale=0.2]{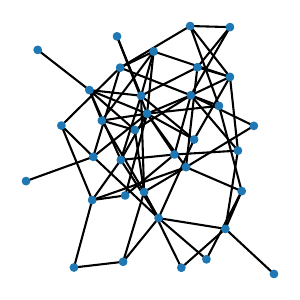}}
         & {\includegraphics[scale=0.2]{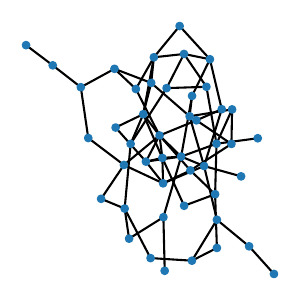}}
         & {\includegraphics[scale=0.2]{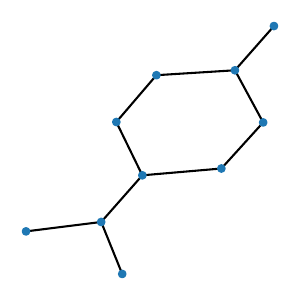}}
         & {\includegraphics[scale=0.2]{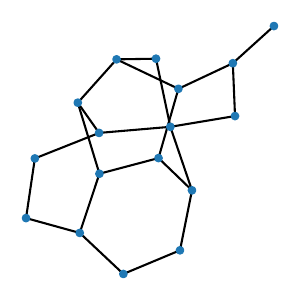}}
         &{\includegraphics[scale=0.2]{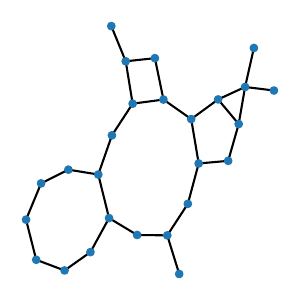}}
         & {\includegraphics[scale=0.2]{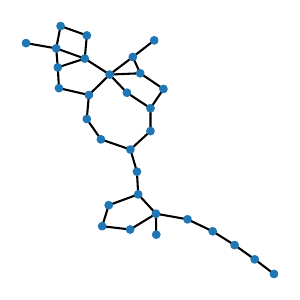}} \\
         
         \begin{turn}{90}\evalmetric\end{turn}
         & 0.09 & 0.00 & 0.24 & 0.10& 0.13 & 0.23 & 0.15 & 0.10 & 0.05 & 0.08\\
         
         \midrule

        % { \begin{turn}{90}DiGress\end{turn} }
        %  & {\includegraphics[scale=0.25]{image/graph_visualization_with_spectral_digress/digress_1102_wordnet_18.32_pred_spec.pdf}}
        %  & {\includegraphics[scale=0.25]{image/graph_visualization_with_spectral_digress/digress_2060_wordnet_4.2_pred_spec.pdf}}
        %  & {\includegraphics[scale=0.25]{image/graph_visualization_with_spectral_digress/digress_21_citeseer_6.91_pred_spec.pdf}}
        %  & {\includegraphics[scale=0.25]{image/graph_visualization_with_spectral_digress/digress_30_citeseer_22.54_pred_spec.pdf}}
        %  & {\includegraphics[scale=0.25]{image/graph_visualization_with_spectral_digress/digress_1350_arxiv_27.76_pred_spec.pdf}}
        %  & {\includegraphics[scale=0.25]{image/graph_visualization_with_spectral_digress/digress_1633_arxiv_29.22_pred_spec.pdf}}
        %  & {\includegraphics[scale=0.25]{image/graph_visualization_with_spectral_digress/digress_7_mutag_7.2_pred_spec.pdf}}
        %  & {\includegraphics[scale=0.25]{image/graph_visualization_with_spectral_digress/digress_5_mutag_9.43_pred_spec.pdf}}
        %  &{\includegraphics[scale=0.25]{image/graph_visualization_with_spectral_digress/digress_20_molbace_11.81_pred_spec.pdf}}
        %  & {\includegraphics[scale=0.25]{image/graph_visualization_with_spectral_digress/digress_35_molbace_14.81_pred_spec.pdf}} \\

        %  \begin{turn}{90}\evalmetric\end{turn}
        %  & 18.32 & 4.2 & 6.91 & 22.54 & 27.76 & 29.22 & 7.20 & 9.43 & 11.81 & 14.81\\

        % \midrule

        { \begin{turn}{90}GruM\end{turn} }
         &  {\includegraphics[scale=0.2]{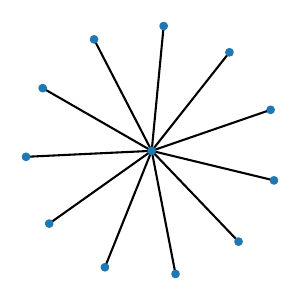}}
         & {\includegraphics[scale=0.2]{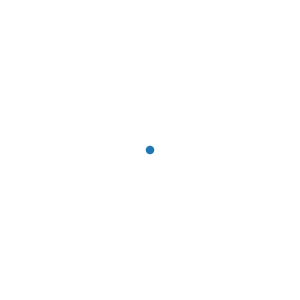}}
         & {\includegraphics[scale=0.2]{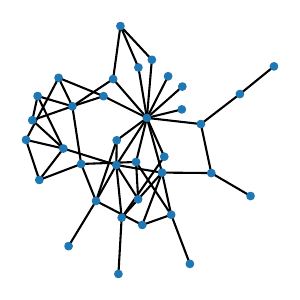}}
         & {\includegraphics[scale=0.2]{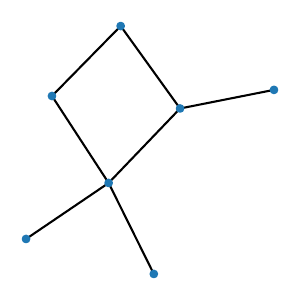}}
         & {\includegraphics[scale=0.2]{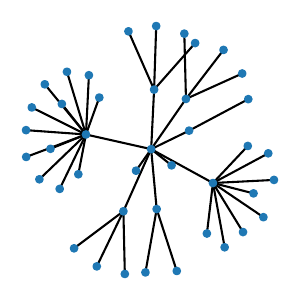}}
         & {\includegraphics[scale=0.2]{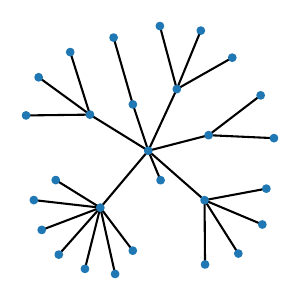}}
         & {\includegraphics[scale=0.2]{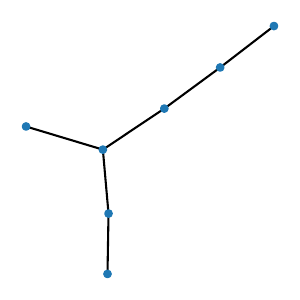}}
         & {\includegraphics[scale=0.2]{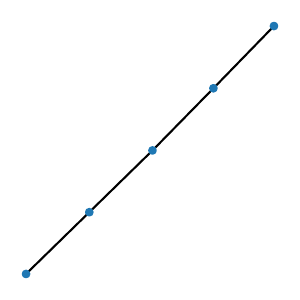}}
         &{\includegraphics[scale=0.2]{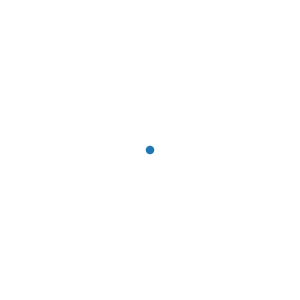}}
         & {\includegraphics[scale=0.2]{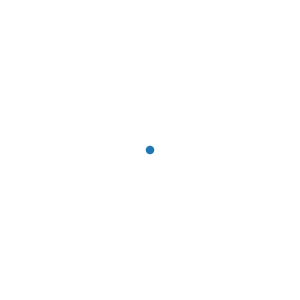}} \\

         \begin{turn}{90}\evalmetric\end{turn}
         & 0.52&  0.97& 0.78& 0.69 & 0.48 & 0.21 & 0.33 & 0.65 & 0.54 & 0.48\\

   \bottomrule
    \end{tabular}
    \label{tab:visual_result}
    \vspace{-10pt}
\end{table*}
\subsection{Model Introspection}
We conduct several ablation studies to understand the effectiveness of \sysname{} in controlled graph generation. We analyze scalability to larger number of nodes; 
provide insights on generating graphs by masking fundamental attributes like number of nodes and edges, while providing all other fine-grained attributes; and 
provide a detailed study on \scheduler{}, to analyze the effects of limiting the inclusion factor and varying the rate of inclusion. 
In addition, we conduct ablation study of \scheduler{} to answer following questions:
(RQ1) Does including the prior distribution \prior{} help?
(RQ2) How does the rate of inclusion affect model's performance?
(RQ3) How much of the prior should be included?

\begin{wrapfigure}[16]{r}{0.45\textwidth}
    % \vspace{-5pt}
    \centering
    \includegraphics[scale=0.4]{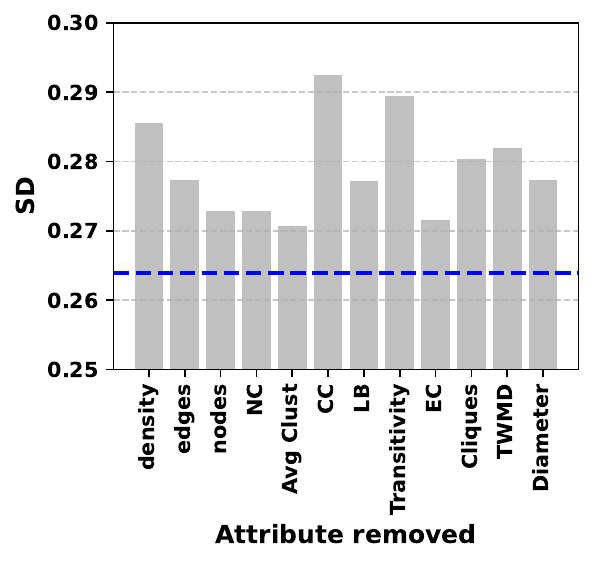}
    % \vspace{-11pt}
    \caption{Plot shows increase in generation error of the specific attribute when not included in training. Blue line indicates including all attributes.}
    \label{fig:citeseer_varying_attribute}
    % \vspace{-21pt}
\end{wrapfigure}

\begin{figure*}[t!]
    \centering
    \subfigure[Performance w/o using number of nodes \& edges as controls.]{\includegraphics[width=0.3\textwidth,height=3cm]{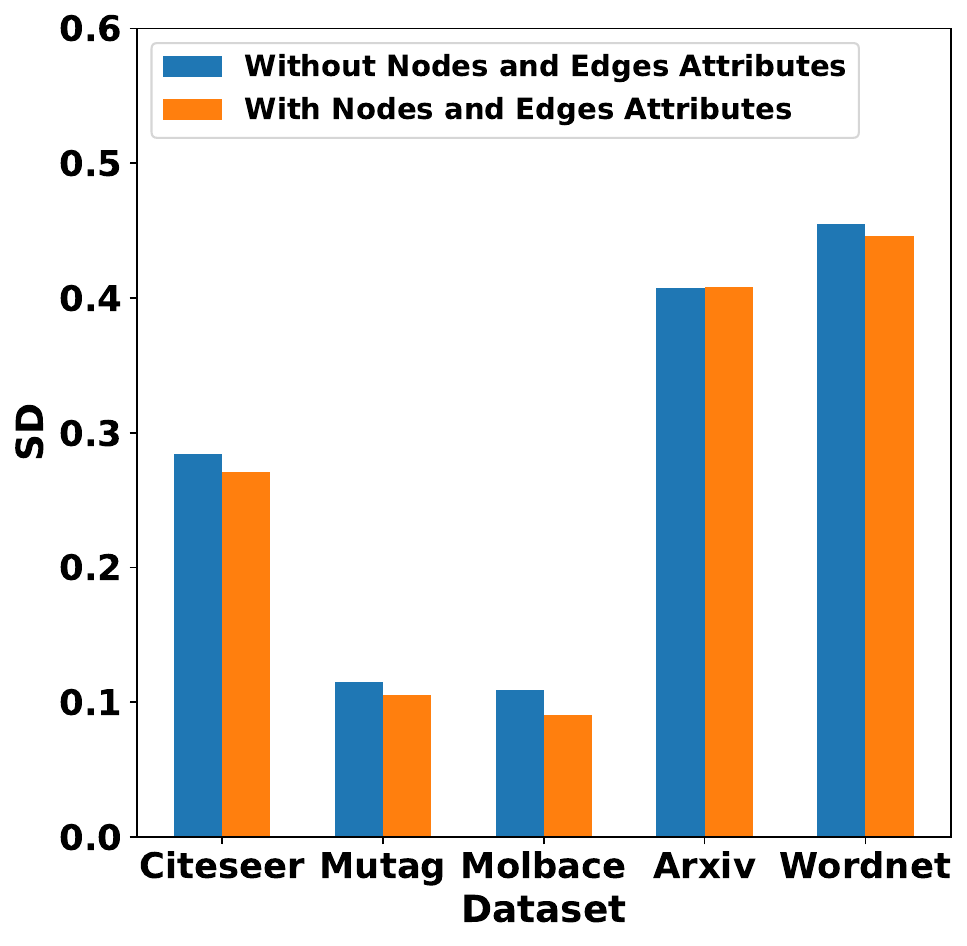}\label{fig:with_and_without_nodes_and_edges}}
    \hfill
    \subfigure[Effect of inclusion factors on generation error across datasets.]{\includegraphics[width=0.3\textwidth,height=3cm]{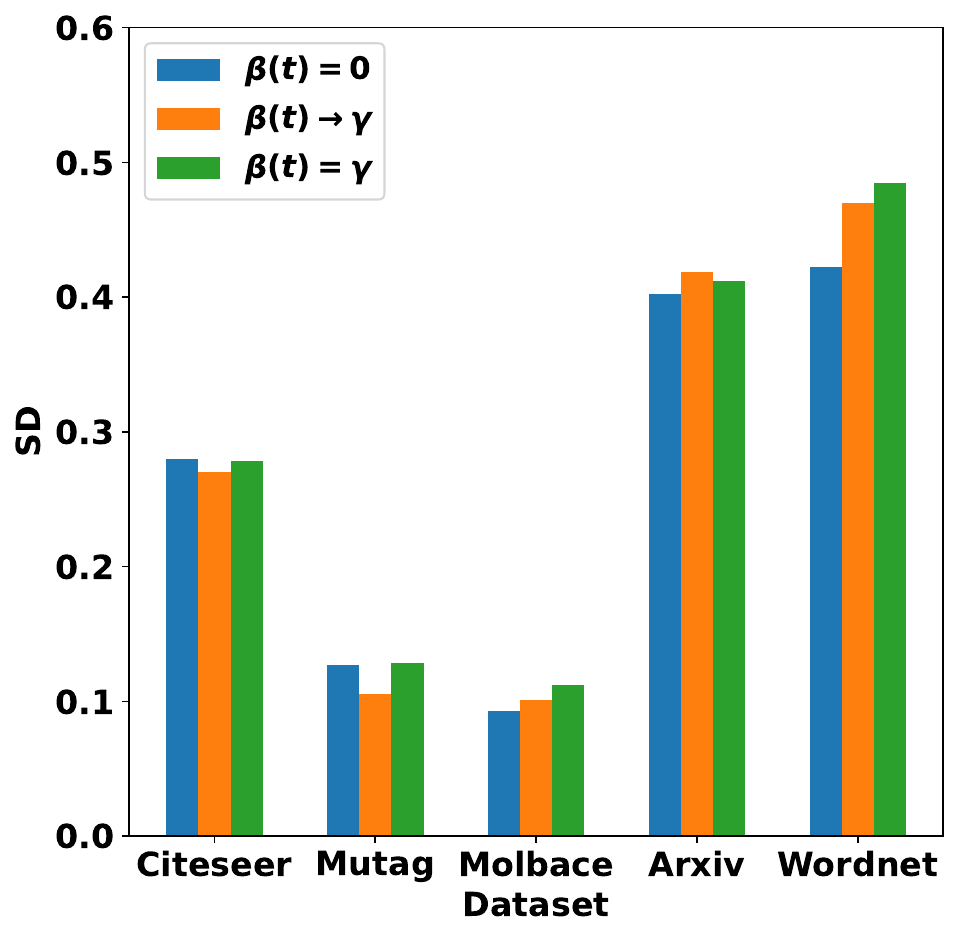}\label{fig:effect_of_beta}} 
    \hfill
    \subfigure[Effect of $\alpha$ on error in generating graphs across datasets.]{\includegraphics[width=0.3\textwidth,height=3cm]{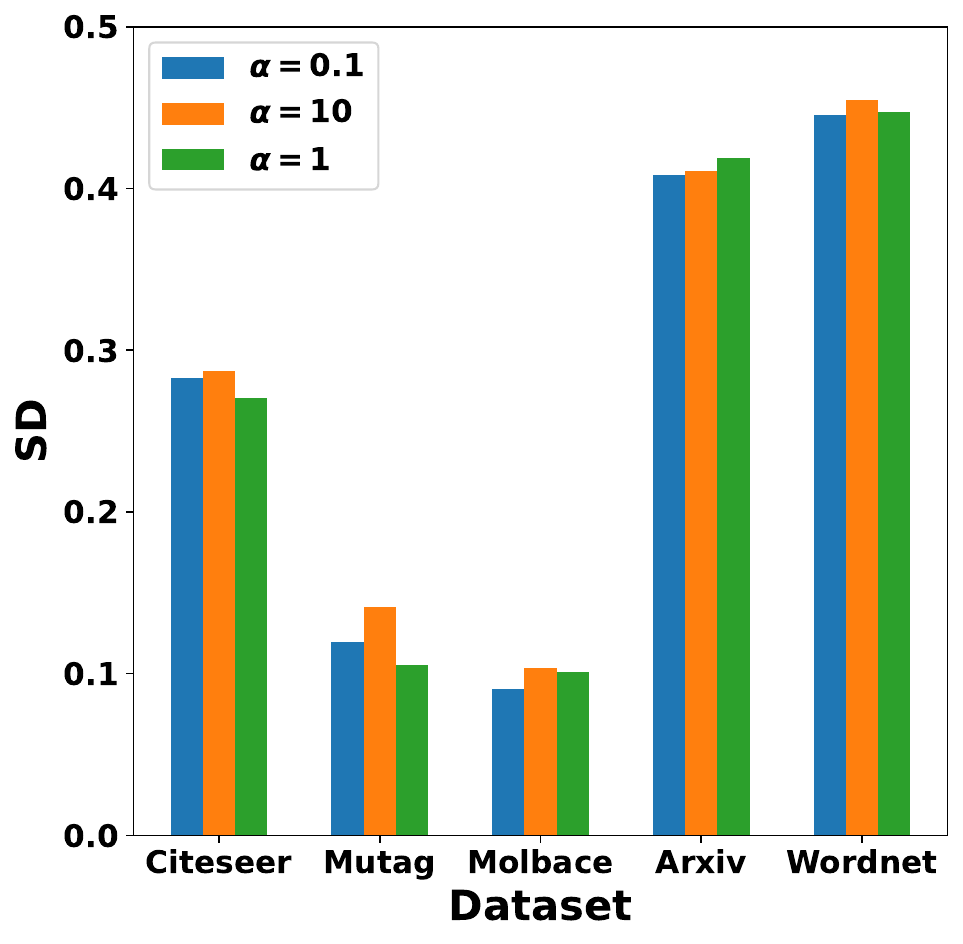}\label{fig:effect_of_alpha}}
    \vspace{-5pt}
    \caption{Ablation Analysis}
    \vspace{-15pt}
\end{figure*}

\vspace{-10pt}
\paragraph{Contribution of control attributes}
Graph attributes determine the required structural properties of generated graphs. Figure~\ref{fig:citeseer_varying_attribute} shows the effect of independently removing one attribute at a time for each training run. Removing either density, closeness centrality or transitivity results in increase in error compared to the number of nodes, average clustering or number of local bridges. This suggests that \sysname{} learns more detailed structural patterns and generates more accurate graphs when guided by a more set of attributes. 
In fact, introducing more restrictive constraints than basic attributes--those such as density or closeness centrality--further refines the generation process and results in graphs that better preserve the intended structural properties. Here, NC (node connectivity), EC (edge connectivity), TWMD (tree width min degree), Avg Clust (average clustering), LB (number of local bridge), Clique (number of cliques).
% Starting with a basic graph attribute (e.g. number of nodes), we retrain the model while adding one randomly selected attribute as additional control for each run, and report the effect on the average performance difference on the original graph attribute (e.g. number of nodes). 
% As Figure~\ref{fig:citeseer_varying_attribute} shows,  increasing the number of control attributes leads to a consistent reduction in error, which eventually stabilizes. This suggests that \sysname{} learns more detailed structural patterns and generates more accurate graphs when guided by a richer set of attributes. In fact, introducing more restrictive constraints than basic attributes--those such as density or closeness centrality--further refines the generation process and results in graphs that better preserve the intended structural properties.
% \begin{wrapfigure}[]{r}{4cm}
%     \vspace{-10pt}
%     \centering
%     \includegraphics[scale=0.27]{image/with_and_without_nodes_and_edges.pdf}
%     \vspace{-10pt}
%     \caption{Comparable performance of \sysname{} with and without using number of nodes and edges as control attributes.}
%     \label{fig:with_and_without_nodes_and_edges}
%     \vspace{-30pt}
% \end{wrapfigure}
\paragraph{Generation without number of nodes and edges}
Figure~\ref{fig:with_and_without_nodes_and_edges} compares the performance of \sysname{} when trained with and without the number of nodes and edges as explicit control attributes. The results show that the model achieves similar performance even without these basic attributes, which suggests that \sysname{} can infer the number of nodes and edges with minimal error using other fine-grained structural attributes. This demonstrates the model's ability to capture graph properties and generate structurally consistent graphs without relying on direct node and edge count supervision, which are commonly used by other models. 
% The performance after removing nodes and edges is slightly increased making it comparable to using all attributes. This indicates that our model \sysname{} unlike any other model can determine the number of nodes and edges with minimum error using rest of the attributes.
\paragraph{Contribution Analysis of components from \sysname{}} 

Table~\ref{tab:obj_ablation} reports an ablation study of the \sysname{} objective function across five datasets highlighting the effective components. In WordNet, removing the \scheduler{} causes a sharp error increase (4.66 vs.\ 0.44), making it the most critical, followed by attribute reconstruction (1.12). In Citeseer, using adjacency matrix increases error to (0.53) and use of GNN as a graph encoder to (0.50), underscoring the role of both attribute learning and type of structural encoding. For Ogbn-Arxiv, GNN as a graph encoder increases the error to (0.57) and the distance function to (1.36) which is the key for aligning prior \prior{} and posterior \posterior{} distributions. In Mutag and Molbace, the distance function (0.49, 0.99) and using adjacency matrices (0.17, 0.17) guide \sysname{} towards lower error. Finally, replacing CNN with a GNN encoder \cite{xu2018how} consistently degrades performance. We hypothesize that this is due to over-smoothing, making GNN struggle to precisely reconstruct the graph structures. These results confirm that each component contributes to the reduction of the generation error, with \scheduler{} and the distance function being the most influential overall.
%, and these results justify the \sysname{} design choices.

% To study the importance of the proposed training paradigm, we performed an ablation study by individually removing the components from the objective function to capture its importance. 
% adjacency matrix encoder and used only attributes as input during training and inference. 
% Table~\ref{tab:obj_ablation} shows the comparison between \sysname{} and \sysname{} with changing the graph encoder to Graph Neural Network (GNN) \cite{xu2018how}, without enforcing the distance function $\mathcal{D}$, without attribute reconstruction, without the mixture scheduler, that is, only using the posterior structural distribution for training. This shows that the generation error increases significantly when we do not use distance function $\mathcal{D}$. 
\begin{table}
\small
\centering
\caption{Effect of removing individual components from the objective function of \sysname{}. Bold indicates the highest error, marking the most influential component.}
% highlighting the most effective component.}
\resizebox{\textwidth}{!}{
\begin{tabular}{lccccc}

        & \textbf{WordNet}& \textbf{Citeseer}& \textbf{Ogbn-Arxiv} & \textbf{MUTAG} &\textbf{MOLBACE}   \\

        \toprule
         & \multicolumn{5}{c}{\textbf{SD}($\downarrow$)} \\
        \midrule

        \sysname{} w GNN as Graph encoder  & 0.32 & 0.50 & 0.57 & 0.18 & 0.15 \\
        \sysname{} w/o Distance Function & 0.69 & 0.45& \textbf{1.36}& \textbf{0.49}&\textbf{0.99} \\
        \sysname{} w/o Attribute Reconstruction &1.12 &0.28 &0.40 &0.14 &0.15 \\
        \sysname{} w/o \scheduler{} & \textbf{4.66} & 0.26 &0.40 & 0.16& 0.13\\
        \sysname{} w/o Adjacency matrices during training & 0.62 &\textbf{0.53} &0.55 &0.17 &0.17 \\
        \midrule
        \textbf{\sysname{}}  &  0.44 &0.27 & 0.40 & 0.10 & 0.09  \\

         \bottomrule
\end{tabular}
}
\label{tab:obj_ablation}
\end{table}
% \begin{table*}[t!]
% \centering
% \caption{Performance of \sysname{} compared with and without using adjacency matrix as input.}
% \resizebox{\textwidth}{!}{
% \begin{tabular}{lccccc|ccccc} 

%         & \textbf{WordNet}& \textbf{Citeseer}& \textbf{Ogbn-Arxiv} & \textbf{MUTAG} &\textbf{MOLBACE}  & 
%         \textbf{WordNet}& \textbf{Citeseer}& \textbf{Ogbn-Arxiv} & \textbf{MUTAG} &\textbf{MOLBACE}  \\

%         \toprule
%          & \multicolumn{5}{c|}{\textbf{\evalmetric}} & \multicolumn{5}{c}{\textbf{GED}}\\
%         \midrule

%         \textbf{\sysname{} w/o graph}    & 0.62  & 0.53 &0.55 & 0.17 & 0.17 &32.06 & 50.02&72.62 & \textbf{12.00}& 34.79 \\

%                 \textbf{\sysname{}}  &  \textbf{0.44} & \textbf{0.27} & \textbf{0.40} & \textbf{0.10} & \textbf{0.09}   & \textbf{26.79}& \textbf{45.62}& \textbf{49.68}& 12.88&\textbf{28.90}  \\

%          \bottomrule
% \end{tabular}
% }
% \label{tab:wo_graph}
% \vspace{-10pt}
% \end{table*}

% \begin{enumerate}[start=1,label={(Q\arabic*):}]
%     \item Does including the prior distribution \prior{} help?
%     \item How does the rate of inclusion affect model's performance?
%     \item How much of the prior should be included?
% \end{enumerate}
\paragraph{RQ1: Does including the prior distribution \prior{} help?} \label{sec:beta_settings}
We consider three scenarios:
(i) when the model only learns from \posterior{} distribution ($\beta(t)$ = 0), 
(ii) when the model gradually combine \prior{} and \posterior{} as training progresses ($\beta(t) \rightarrow \gamma$), and 
(iii) when the model combines both \prior{} and \posterior{}  with constant influence factor $\beta(t) = \gamma$. 
As shown in Figure \ref{fig:effect_of_beta}, combining representations from both distributions \prior{} and \posterior{} helps generate better graphs compared to using only representations from \posterior{}. Also, gradual increase in influence factor $\beta(t)\rightarrow\gamma$ performs better compared to keeping it  constant $\beta(t)$. We conclude relying only on graph representation from \posterior{} without considering attribute representation from \prior{}  results in higher \evalmetric{} error and lower performance. 

% \begin{figure*}[t!]
%     \centering
%     \subfigure[Relation between the maximum inclusion rate $\gamma$ and performance (\evalmetric{} errors). \scheduler{} reduces \evalmetric{} error by combining information form both distributions]{\includegraphics[width=0.4\textwidth]{image/percent_sample_p_vs_spectral_norm.png}\label{fig:p_percent_sample}}
%     \hfill
%     \subfigure[\evalmetric{} on test data when masking only one attribute with zero while keeping others unchanged. The dotted line shows \sysname{}'s performance on Citeseer without any masking. Each bar shows \evalmetric{} when a specific attribute is masked.]{\includegraphics[width=0.4\textwidth]{image/denoising_Attributes_spectral_norm.pdf}\label{fig:denoise_attribute}} 
%     \caption{Ablation Analysis}   
% \end{figure*}

\begin{wrapfigure}[15]{r}{0.45\textwidth}
        \vspace{-15pt}
        \centering
        \includegraphics[scale=0.25]{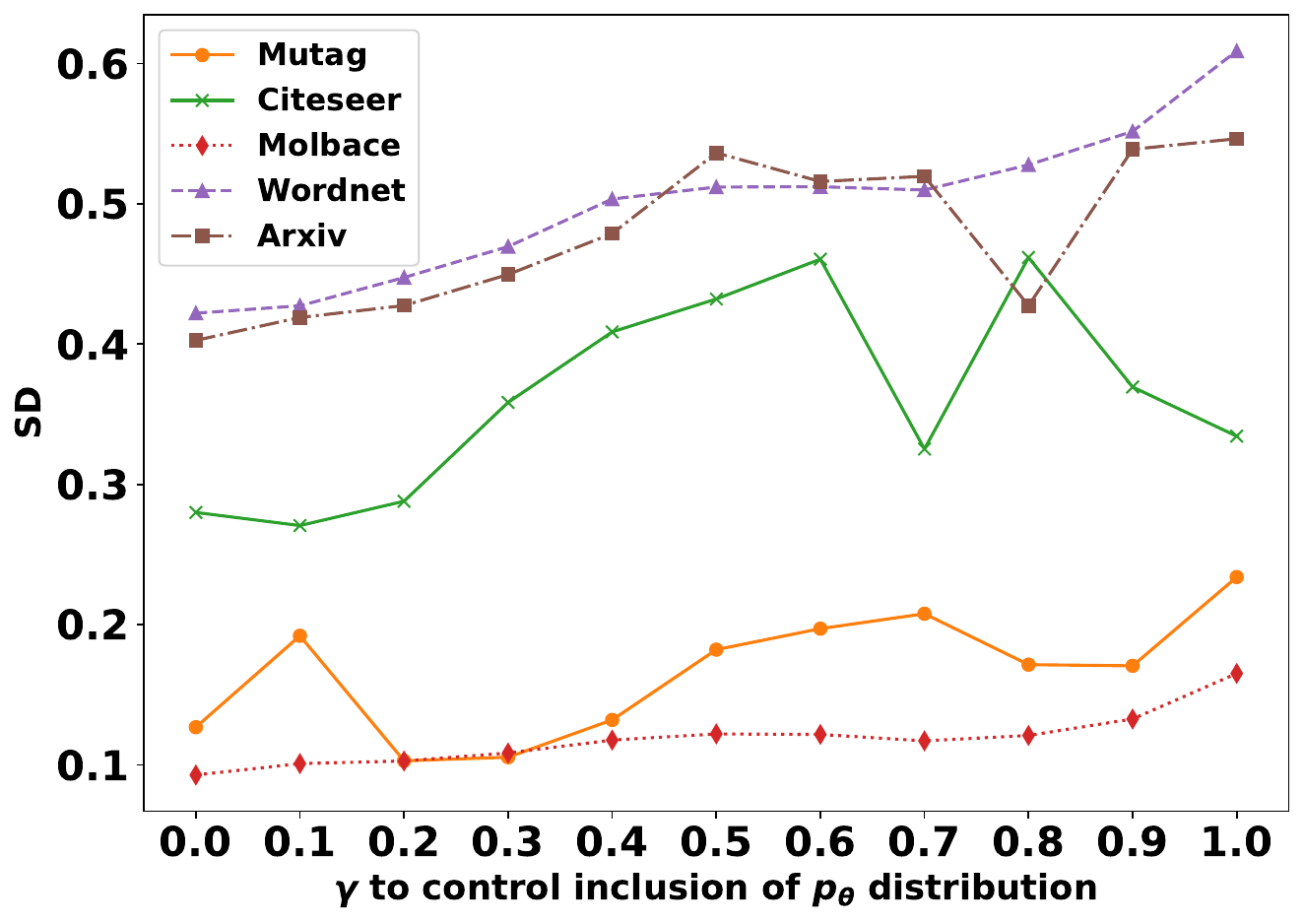}
        \vspace{-10pt}
        \caption{Relation between the maximum inclusion rate $\gamma$ and  \evalmetric{} error. \scheduler{} reduces \evalmetric{} error by combining information from both distributions.\looseness-1}
        \label{fig:p_percent_sample}
        % \vspace{-20pt} 
\end{wrapfigure}
\paragraph{RQ2:  How does the rate of inclusion affect model's performance?}
\label{sec:effect_of_alpha} 
We analyze different rates of inclusion. As Figure~\ref{fig:effect_of_alpha} shows, a slow inclusion rate ($\alpha$ = 0.1)  often helps model in learning better representations compared faster inclusion rates, e.g. ($\alpha$=10). This result suggests that initially focusing on the \posterior{} and gradually incorporating the \prior{} yields better latent representations.\looseness-1
\paragraph{RQ3: How much of the prior should be included?} \label{sec:gamma_threshold}
We vary the influence of prior distribution by adjusting the maximum inclusion rate, $\gamma\in [0,1]$, % with step size of 0.1. 
where $\gamma$ =0 excludes \prior{} entirely, and $\gamma$ = 1 excludes the posterior \posterior{}. Figure~\ref{fig:p_percent_sample} shows that smaller values of $\gamma$ result in lower error, suggesting that limited inclusion of \prior{} improves graph generation by better balancing both distributions.\looseness-1
\subsection{De-noising graph attributes}
\begin{wrapfigure}[10]{r}{0.45\textwidth}
\vspace{-70pt}
    \centering
        \includegraphics[scale=0.30]{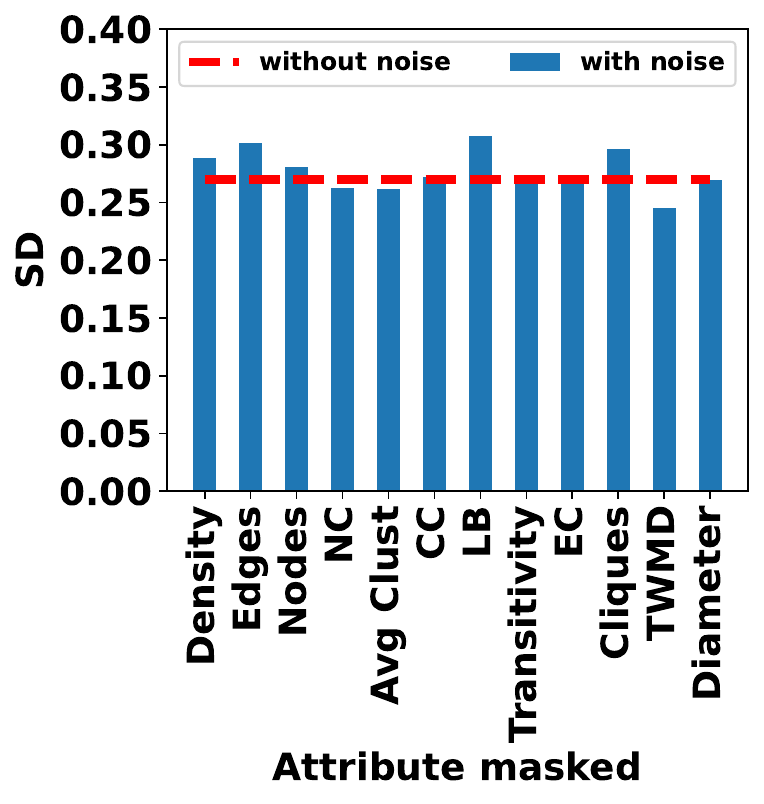}
        \vspace{-11pt}
        % \caption{\evalmetric{} on test data when masking only one attribute with zero while keeping others unchanged. The dotted line shows \sysname{}'s performance on Citeseer without any masking. Each bar shows \evalmetric{} when a specific attribute is masked.}
        \caption{Bars indicate \evalmetric{} on Citeseer with one attribute masked (set to zero) at a time. The dotted line marks \sysname{}’s performance without masking.}
        \label{fig:denoise_attribute}
\end{wrapfigure}
% We evaluate \sysname{}'s robustness to noisy attributes by masking one attribute at a time during inference. For this experiment, we use our best trained model and freeze its parameters, and run inference 12 times, once per attribute making its value to zero with other unchanged on entire test graphs. Figure~\ref{fig:denoise_attribute} shows the results, where the dotted horizontal line shows the \evalmetric{} error value of \sysname{} without masking any attribute and serves as a reference. The results show that \sysname{} is often able to ignore noisy attributes and generate accurate graphs based on the remaining attributes, which demonstrates its resilience to missing control attributes. 
% In addition, the highest SD values occur when edges, local bridge or cliques are masked, indicating that these attributes are critical for structural fidelity. Attributes such as clustering coefficient, transitivity or diameter show moderate or no increases in SD when masked, indicating that these attributes contribute to finer structural properties but are not as fundamental as others.
We evaluate \sysname{}’s robustness to noisy attributes by masking one attribute at a time during inference. Using the best trained model with frozen parameters, we run 12 inference passes, each time setting one attribute to zero across all test graphs. Figure~\ref{fig:denoise_attribute} shows the results, with the dotted line as the baseline \evalmetric{} error without masking.\sysname{} remains resilient, often generating accurate graphs despite missing controls. The largest error increases occur when edges, local bridges, or cliques are masked, confirming their critical role in structural fidelity, whereas masking clustering coefficient, transitivity, or diameter yields only minor changes, indicating they refine finer structural details.

% uncomment to go back to kdd format 
% \begin{figure}
%      \centering
%         \includegraphics[scale=.35]{image/denoising_Attributes_spectral.pdf}
%         \vspace{-10pt}
%         \caption{\evalmetric{} on test data when masking only one attribute with zero while keeping others unchanged. The dotted line shows \sysname{}'s performance on Citeseer without any masking. Each bar shows \evalmetric{} when a specific attribute is masked. Abbreviations NC (node connectivity), EC (edge connectivity), TWMD (tree width min degree), Avg Clust (average clustering), LB (number of local bridge), Clique (number of cliques).}
%         \label{fig:denoise_attribute}
%     \vspace{-10pt}
% \end{figure}

% \subsection{Effect of changing Loss function}
% KL loss vs. JS Loss vs. WLD Loss

\section{Conclusion and Future Work}
% We introduce a novel approach to controlled graph generation, named \sysname{}, which can generate graphs that satisfy given topological attributes. The model learns the latent representations of the graph and attributes, and effectively integrates using a scheduling technique \scheduler{}. \scheduler{} combines the attribute representations from the prior and graph representations from the posterior to produce final latent representations for graph generator. In future, we will extend our approach to generate desired graphs using effective training techniques with minimal training data.

We presented \sysname{}, a novel controlled graph generation model that generates graphs satisfying fine-grained topological attributes. It includes a novel distribution scheduler, \scheduler{}, to combines {\em attribute} and {\em adjacency matrix} representations for learning accurate latent structures. \sysname{} enables precise control--even without explicitly specifying basic properties such as node and edge counts--and achieves lower generation error by gradually integrating multiple control attributes. In future, we plan to extend \sysname{} to dynamic or temporal graphs for applications such as in social network analysis, traffic prediction, and temporal knowledge graphs.

\bibliographystyle{plainnat}
\bibliography{iclr2026}

@ARTICLE{sanfeliu1983distance,
  author={Sanfeliu, Alberto and Fu, King-Sun},
  journal={IEEE Transactions on Systems, Man, and Cybernetics}, 
  title={A distance measure between attributed relational graphs for pattern recognition}, 
  year={1983},
  volume={SMC-13},
  number={3},
  pages={353-362},
  doi={10.1109/TSMC.1983.6313167}}

@article{zeng2009comparing,
  title={Comparing stars: On approximating graph edit distance},
  author={Zeng, Zhiping and Tung, Anthony KH and Wang, Jianyong and Feng, Jianhua and Zhou, Lizhu},
  journal={Proceedings of the VLDB Endowment},
  volume={2},
  number={1},
  pages={25--36},
  year={2009},
  publisher={VLDB Endowment}
}

@article{miller1995wordnet,
  title={WordNet: a lexical database for English},
  author={Miller, George A},
  url={https://dl.acm.org/doi/pdf/10.1145/219717.219748},
  journal={Communications of the ACM},
  volume={38},
  number={11},
  pages={39--41},
  year={1995},
  publisher={ACM New York, NY, USA}
}

@article{hu2020open,
  title={Open graph benchmark: Datasets for machine learning on graphs},
  author={Hu, Weihua and Fey, Matthias and Zitnik, Marinka and Dong, Yuxiao and Ren, Hongyu and Liu, Bowen and Catasta, Michele and Leskovec, Jure},
  url={https://proceedings.neurips.cc/paper/2020/hash/fb60d411a5c5b72b2e7d3527cfc84fd0-Abstract.html},
  journal={Advances in neural information processing systems},
  volume={33},
  pages={22118--22133},
  year={2020}
}

@inproceedings{kipfW17,
  author    = {Thomas N. Kipf and
               Max Welling},
  title     = {Semi-Supervised Classification with Graph Convolutional Networks},
  booktitle = {5th International Conference on Learning Representations, {ICLR} 2017,
               Toulon, France, April 24-26, 2017, Conference Track Proceedings},
  publisher = {OpenReview.net},
  year      = {2017},
  url       = {https://openreview.net/forum?id=SJU4ayYgl},
  bibsource = {dblp computer science bibliography, https://dblp.org}
}

@article{DBLP:journals/corr/abs-2007-08663,
  author       = {Christopher Morris and
                  Nils M. Kriege and
                  Franka Bause and
                  Kristian Kersting and
                  Petra Mutzel and
                  Marion Neumann},
  title        = {TUDataset: {A} collection of benchmark datasets for learning with
                  graphs},
  journal      = {CoRR},
  volume       = {abs/2007.08663},
  year         = {2020},
  url          = {https://arxiv.org/abs/2007.08663},
  eprinttype    = {arXiv},
  eprint       = {2007.08663},
  bibsource    = {dblp computer science bibliography, https://dblp.org}
}

@book{pitas2016graph,
  title={Graph-based social media analysis},
  author={Pitas, Ioannis},
  year={2016},
  publisher={CRC Press}
}

@InProceedings{pmlr-v80-you18a,
  title = 	 {{G}raph{RNN}: Generating Realistic Graphs with Deep Auto-regressive Models},
  author =       {You, Jiaxuan and Ying, Rex and Ren, Xiang and Hamilton, William and Leskovec, Jure},
  booktitle = 	 {Proceedings of the 35th International Conference on Machine Learning},
  pages = 	 {5708--5717},
  year = 	 {2018},
  editor = 	 {Dy, Jennifer and Krause, Andreas},
  volume = 	 {80},
  series = 	 {Proceedings of Machine Learning Research},
  month = 	 {10--15 Jul},
  publisher =    {PMLR},
  url = 	 {https://proceedings.mlr.press/v80/you18a.html}
}

@inproceedings{jin2020hierarchical,
  title={Hierarchical generation of molecular graphs using structural motifs},
  author={Jin, Wengong and Barzilay, Regina and Jaakkola, Tommi},
  booktitle={International conference on machine learning},
  pages={4839--4848},
  year={2020},
  url={https://proceedings.mlr.press/v119/jin20a/jin20a.pdf},
  organization={PMLR}
}

@inproceedings{jin2018junction,
  title={Junction tree variational autoencoder for molecular graph generation},
  author={Jin, Wengong and Barzilay, Regina and Jaakkola, Tommi},
  booktitle={International conference on machine learning},
  pages={2323--2332},
  year={2018},
  url ={http://proceedings.mlr.press/v80/jin18a/jin18a.pdf},
  organization={PMLR}
}

@article{wang2022deep,
  title={Deep generative model for periodic graphs},
  author={Wang, Shiyu and Guo, Xiaojie and Zhao, Liang},
  journal={Advances in Neural Information Processing Systems},
  volume={35},
  url = {https://proceedings.neurips.cc/paper_files/paper/2022/file/e89e8f84626197942b36a82e524c2529-Paper-Conference.pdf},
  year={2022}
}

@article{Erdos:1959:pmd,
  title = {2017-10-20T13:47:06.000+0200},
  author = {Erd\"os, P and R\'enyi, A},
  url = {https://publi.math.unideb.hu/load_doi.php?pdoi=10_5486_PMD_1959_6_3_4_12},
  journal = {Publicationes Mathematicae Debrecen},
  pages = {290--297},
  title = {On Random Graphs I},
  volume = 6,
  year = 1959
}

@article{barabasi1999emergence,
  title={Emergence of scaling in random networks},
  author={Barab{\'a}si, Albert-L{\'a}szl{\'o} and Albert, R{\'e}ka},
  url={https://www.science.org/doi/full/10.1126/science.286.5439.509},
  journal={science},
  volume={286},
  number={5439},
  pages={509--512},
  year={1999},
  publisher={American Association for the Advancement of Science}
}

@inproceedings{melnyk-etal-2022-knowledge,
    title = "Knowledge Graph Generation From Text",
    author = "Melnyk, Igor  and
      Dognin, Pierre  and
      Das, Payel",
    editor = "Goldberg, Yoav  and
      Kozareva, Zornitsa  and
      Zhang, Yue",
    booktitle = "Findings of the Association for Computational Linguistics: EMNLP 2022",
    month = dec,
    year = "2022",
    address = "Abu Dhabi, United Arab Emirates",
    publisher = "Association for Computational Linguistics",
    url = "https://aclanthology.org/2022.findings-emnlp.116",
    doi = "10.18653/v1/2022.findings-emnlp.116",
    pages = "1610--1622",
}

@inproceedings{luo2021graphdf,
  title={Graphdf: A discrete flow model for molecular graph generation},
  author={Luo, Youzhi and Yan, Keqiang and Ji, Shuiwang},
  booktitle={International Conference on Machine Learning},
  pages={7192--7203},
  year={2021},
  url = {https://proceedings.mlr.press/v139/luo21a.html},
  organization={PMLR}
}

@article{popova2019molecularrnn,
  title={MolecularRNN: Generating realistic molecular graphs with optimized properties},
  author={Popova, Mariya and Shvets, Mykhailo and Oliva, Junier and Isayev, Olexandr},
  journal={arXiv preprint arXiv:1905.13372},
  url = {https://arxiv.org/pdf/1905.13372},
  year={2019}
}

@inproceedings{shi2019graphaf,
  title={GraphAF: a Flow-based Autoregressive Model for Molecular Graph Generation},
  author={Shi, Chence and Xu, Minkai and Zhu, Zhaocheng and Zhang, Weinan and Zhang, Ming and Tang, Jian},
  booktitle={International Conference on Learning Representations},
  url={https://openreview.net/pdf?id=S1esMkHYPr},
  year={2019}
}

@inproceedings{DBLP:conf/emnlp/CaoH0LXLJZ23,
  author       = {Pengfei Cao and
                  Yupu Hao and
                  Yubo Chen and
                  Kang Liu and
                  Jiexin Xu and
                  Huaijun Li and
                  Xiaojian Jiang and
                  Jun Zhao},
  editor       = {Houda Bouamor and
                  Juan Pino and
                  Kalika Bali},
  title        = {Event Ontology Completion with Hierarchical Structure Evolution Networks},
  booktitle    = {Proceedings of the 2023 Conference on Empirical Methods in Natural
                  Language Processing, {EMNLP} 2023, Singapore, December 6-10, 2023},
  pages        = {306--320},
  publisher    = {Association for Computational Linguistics},
  year         = {2023},
  url          = {https://aclanthology.org/2023.emnlp-main.21},
  bibsource    = {dblp computer science bibliography, https://dblp.org}
}

@inproceedings{DBLP:conf/emnlp/ZhouZG0023,
  author       = {Wentao Zhou and
                  Jun Zhao and
                  Tao Gui and
                  Qi Zhang and
                  Xuanjing Huang},
  editor       = {Houda Bouamor and
                  Juan Pino and
                  Kalika Bali},
  title        = {Inductive Relation Inference of Knowledge Graph Enhanced by Ontology
                  Information},
  booktitle    = {Findings of the Association for Computational Linguistics: {EMNLP}
                  2023, Singapore, December 6-10, 2023},
  pages        = {6491--6502},
  publisher    = {Association for Computational Linguistics},
  year         = {2023},
  url          = {https://aclanthology.org/2023.findings-emnlp.431},
  bibsource    = {dblp computer science bibliography, https://dblp.org}
}

@inproceedings{allamanis2018learning,
  title={Learning to Represent Programs with Graphs},
  author={Allamanis, Miltiadis and Brockschmidt, Marc and Khademi, Mahmoud},
  booktitle={International Conference on Learning Representations},
  year={2018},
  url = {https://openreview.net/pdf?id=BJOFETxR-}
}

@techreport{hagberg2008exploring,
  title={Exploring network structure, dynamics, and function using NetworkX},
  author={Hagberg, Aric and Swart, Pieter and S Chult, Daniel},
  year={2008},
  institution={Los Alamos National Lab.(LANL), Los Alamos, NM (United States)},
 url = {https://www.researchgate.net/publication/236407765_Exploring_Network_Structure_Dynamics_and_Function_Using_NetworkX}
}

@article{zahirnia2024neural,
  title={Neural Graph Generation from Graph Statistics},
  author={Zahirnia, Kiarash and Hu, Yaochen and Coates, Mark and Schulte, Oliver},
  journal={Advances in Neural Information Processing Systems},
  url={https://proceedings.neurips.cc/paper_files/paper/2023/file/72153267883fbcafdb6e4662382696c5-Paper-Conference.pdf},
  volume={36},
  year={2024}
}

@inproceedings{chen2023efficient,
  title={Efficient and Degree-Guided Graph Generation via Discrete Diffusion Modeling},
  author={Chen, Xiaohui and He, Jiaxing and Han, Xu and Liu, Liping},
  booktitle={International Conference on Machine Learning},
  url={https://proceedings.mlr.press/v202/chen23k/chen23k.pdf},
  pages={4585--4610},
  year={2023},
  organization={PMLR}
}

@inproceedings{martinkus2022spectre,
  title={Spectre: Spectral conditioning helps to overcome the expressivity limits of one-shot graph generators},
  author={Martinkus, Karolis and Loukas, Andreas and Perraudin, Nathana{\"e}l and Wattenhofer, Roger},
  booktitle={International Conference on Machine Learning},
  url = {https://proceedings.mlr.press/v162/martinkus22a/martinkus22a.pdf},
  pages={15159--15179},
  year={2022},
  organization={PMLR}
}

@article{kullback1951information,
  title={On information and sufficiency},
  author={Kullback, S and Leibler, RA},
  journal={The annals of mathematical statistics},
  volume={22},
  number={1},
  pages={79--86},
  year={1951},
  publisher={JSTOR},
  url = {https://www.jstor.org/stable/2236703}
}

@article{d2eb0123,
 ISSN = {00251909, 15265501},
 URL = {http://www.jstor.org/stable/2627082},
 author = {L. V. Kantorovich},
 journal = {Management Science},
 number = {4},
 pages = {366--422},
 publisher = {INFORMS},
 title = {Mathematical Methods of Organizing and Planning Production},
 urldate = {2024-10-01},
 volume = {6},
 year = {1960}
}

@article{de2018molgan,
  title={{MolGAN: An implicit generative model for small
  molecular graphs}},
  author={De Cao, Nicola and Kipf, Thomas},
  journal={ICML 2018 workshop on Theoretical Foundations 
  and Applications of Deep Generative Models},
  url={https://arxiv.org/pdf/1805.11973},
  year={2018}
}

@inproceedings{zang2020moflow,
  title={Moflow: an invertible flow model for generating molecular graphs},
  author={Zang, Chengxi and Wang, Fei},
  booktitle={Proceedings of the 26th ACM SIGKDD international conference on knowledge discovery \& data mining},
  pages={617--626},
  year={2020},
  url = {https://dl.acm.org/doi/pdf/10.1145/3394486.3403104}
}

@inproceedings{
liu2021graphebm,
title={Graph{EBM}: Molecular Graph Generation with Energy-Based Models},
author={Meng Liu and Keqiang Yan and Bora Oztekin and Shuiwang Ji},
booktitle={Energy Based Models Workshop - ICLR 2021},
year={2021},
url={https://openreview.net/forum?id=Gc51PtL_zYw}
}

@inproceedings{
Shi2020GraphAF,
title={GraphAF: a Flow-based Autoregressive Model for Molecular Graph Generation},
author={Chence Shi and Minkai Xu and Zhaocheng Zhu and Weinan Zhang and Ming Zhang and Jian Tang},
booktitle={International Conference on Learning Representations},
year={2020},
url={https://openreview.net/forum?id=S1esMkHYPr}
}

@article{sanchez2018inverse,
  title={Inverse molecular design using machine learning: Generative models for matter engineering},
  author={Sanchez-Lengeling, Benjamin and Aspuru-Guzik, Al{\'a}n},
  journal={Science},
  volume={361},
  number={6400},
  pages={360--365},
  year={2018},
  publisher={American Association for the Advancement of Science},
  url = {https://www.science.org/doi/10.1126/science.aat2663}
}

@inproceedings{zeno2021dymond,
  title={Dymond: Dynamic motif-nodes network generative model},
  author={Zeno, Giselle and La Fond, Timothy and Neville, Jennifer},
  booktitle={Proceedings of the Web Conference 2021},
  pages={718--729},
  year={2021},
  url = {https://arxiv.org/pdf/2308.00770}
}

@inproceedings{zhou2020data,
  title={A data-driven graph generative model for temporal interaction networks},
  author={Zhou, Dawei and Zheng, Lecheng and Han, Jiawei and He, Jingrui},
  booktitle={Proceedings of the 26th ACM SIGKDD International Conference on Knowledge Discovery \& Data Mining},
  pages={401--411},
  year={2020},
  url = {https://dl.acm.org/doi/pdf/10.1145/3394486.3403082}
}

@inproceedings{vignacdigress,
  title={DiGress: Discrete Denoising diffusion for graph generation},
  author={Vignac, Clement and Krawczuk, Igor and Siraudin, Antoine and Wang, Bohan and Cevher, Volkan and Frossard, Pascal},
  booktitle={The Eleventh International Conference on Learning Representations},
  year = {2023},
   url = {https://openreview.net/pdf?id=UaAD-Nu86WX},
   
}

@inproceedings{jograph,
  title={Graph Generation with Diffusion Mixture},
  author={Jo, Jaehyeong and Kim, Dongki and Hwang, Sung Ju},
  booktitle={Forty-first International Conference on Machine Learning},
  year = {2024},
  url = {https://openreview.net/pdf?id=cZTFxktg23s}
}

@article{yang2019conditional,
  title={Conditional structure generation through graph variational generative adversarial nets},
  author={Yang, Carl and Zhuang, Peiye and Shi, Wenhan and Luu, Alan and Li, Pan},
  journal={Advances in neural information processing systems},
  volume={32},
  year={2019}
}

@inproceedings{ommi2022ccgg,
  title={Ccgg: A deep autoregressive model for class-conditional graph generation},
  author={Ommi, Yassaman and Yousefabadi, Matin and Faez, Faezeh and Sabour, Amirmojtaba and Soleymani Baghshah, Mahdieh and Rabiee, Hamid R},
  booktitle={Companion Proceedings of the Web Conference 2022},
  pages={1092--1098},
  year={2022}
}

@inproceedings{xu2018how,
  title        = {How Powerful Are Graph Neural Networks?},
  author       = {Keyulu Xu and Weihua Hu and Jure Leskovec and Stefanie Jegelka},
  booktitle    = {International Conference on Learning Representations (ICLR)},
  year         = {2019},
  arXiv        = {1810.00826},
}

@book{Murphy2012,
  author    = {Kevin P. Murphy},
  title     = {Machine Learning: A Probabilistic Perspective},
  year      = {2012},
  publisher = {MIT Press}
}

@article{liu2024graph,
  title={Graph diffusion transformers for multi-conditional molecular generation},
  author={Liu, Gang and Xu, Jiaxin and Luo, Tengfei and Jiang, Meng},
  journal={Advances in Neural Information Processing Systems},
  volume={37},
  pages={8065--8092},
  year={2024}
}

@article{mercatali2024diffusion,
  title={Diffusion twigs with loop guidance for conditional graph generation},
  author={Mercatali, Giangiacomo and Verma, Yogesh and Freitas, Andre and Garg, Vikas},
  journal={Advances in Neural Information Processing Systems},
  volume={37},
  pages={137741--137767},
  year={2024}
}

\section{Appendix}

% \section{Technical Appendices and Supplementary Material}

\subsection{Scalability to fine-grained conditionally generate large graphs}

We analyze the effect of increasing the maximum number of nodes, $|V|$, on \sysname{}'s \evalmetric{} performance. Table~\ref{tab:max_num_node_study} shows that \evalmetric{} increases as the maximum number of nodes grows up to 200 nodes. This is because larger graphs have greater structural complexity, with more potential edges and relationships that are harder to generate accurately. This makes it challenging for the model to capture both local and global topological properties, and potentially leads to cumulative errors in matching node-specific attributes such as degrees and centrality. In addition, larger graphs often contain more variability and sparsity, which further complicates satisfying the desired structural attributes and result in higher deviations between the generated and target graphs.
\begin{table}[h!]
    \small
    \centering
    % \vspace{-18pt}
    \caption{Generation performance degrades as the target number of nodes increases.}
    \begin{tabular}{c c }
        \textbf{\#Nodes} & \textbf{\evalmetric{}}($\downarrow$) \\
        \toprule
        50 & 0.40\\
        60 & 0.60\\
        80 & 0.74 \\
        100 & 0.82\\
        200 & 0.83\\
        \bottomrule
    \end{tabular}
    % \vspace{-10pt}
    \label{tab:max_num_node_study}
\end{table}

\subsection{Limitation}
\sysname{} performs significantly well compared to recent baselines to generate graphs from given fine-grained control attributes. However, the performance degrades as we aim to generate  larger graphs because of two factors: increased complexity in aligning multiple fine-grained attribute constraints as graph size grows, and the use of less expressive encoders (such as convolutional encoders) for capturing long-range dependencies in large graphs. 
\subsection{Settings}
Following previous works~\citep{de2018molgan,zahirnia2024neural}, we set the maximum number of nodes to $V=50$ in experiments. This threshold is appropriate, given the common practice of sampling 1-2 hop subgraphs for nodes. 
% for GNNs due to the nature of how GNNs process graph data,
We set the number of hops to $k=2$ for all datasets except for Citeseer, for which we use $k=3$ due to its smaller size. In addition, we extract graph attributes using Networkx~\citep{hagberg2008exploring}. We consider a maximum number of 1000 training iterations for Citeseer and 200 iterations for other datasets, which is sufficiently large for convergence. 
For the CNN encoder, we use two convolutional layers with kernel size of 5, and 32 and 64 channels respectively for all datasets.  For the decoder, we used two convolutional layers with 64,32 channels respectively. 
blueThe model requires approximately 260M FLOPs for the graph encoder, 260M for the graph decoder, 231M for the attribute encoder, and 231M for the attribute decoder. In total, it contains about 116M trainable parameters, consisting of roughly 100k from the CNN components and 115M from the MLP with a hidden dimension size of 1024.
For hyperparameters, we set the maximum possible inclusion from prior \prior{} ($\gamma$) to 0.3 for Mutag, 0.1 for Molbace, Citeseer, and arxiv; and 0.2 for Wordnet.
We consider a batch-size of 1,028 and run all our experiments on a single A100 40GB GPU.

\subsection{Graph Visualization}\label{app:vis}
Table~\ref{tab:other_visual_result} shows examples of different graphs generated by baselines across datasets.

\begin{table*}[h!]
    \small
    % \tiny
    \centering
    \caption{Graph visualization across datasets. Examples are taken from test splits of datasets. }
    \begin{tabular}{p{0.2cm} p{0.9cm} p{0.9cm} p{0.9cm} p{0.9cm} p{0.9cm} p{0.9cm} p{0.9cm} p{0.9cm} p{0.9cm} p{0.9cm}}
         & \multicolumn{2}{c}{\textbf{Wordnet}} & \multicolumn{2}{c}{\textbf{Citeseer}} & \multicolumn{2}{c}{\textbf{Ogbn-Arxiv}} & \multicolumn{2}{c}{\textbf{Mutag}} & \multicolumn{2}{c}{\textbf{Molbace}}\\
         \toprule
         
         {\begin{turn}{90} Test \end{turn}} 
         & {\includegraphics[scale=0.25]{image/graph_visualization_with_mad/1102_wordnet_0.85_gt.pdf}}  
         & {\includegraphics[scale=0.25]{image/graph_visualization_with_mad/2060_wordnet_0.0_gt.pdf}}
         & {\includegraphics[scale=0.25]{image/graph_visualization_with_mad/21_citeseer_0.1_gt.pdf}}
         & {\includegraphics[scale=0.25]{image/graph_visualization_with_mad/30_citeseer_2.36_gt.pdf}}
         & {\includegraphics[scale=0.25]{image/graph_visualization_with_mad/1350_arxiv_2.85_gt.pdf}}
         & {\includegraphics[scale=0.25]{image/graph_visualization_with_mad/1633_arxiv_3.71_gt.pdf}}
         & {\includegraphics[scale=0.25]{image/graph_visualization_with_mad/7_mutag_0.76_gt.pdf}}
         & {\includegraphics[scale=0.25]{image/graph_visualization_with_mad/5_mutag_1.09_gt.pdf}}
         &{\includegraphics[scale=0.25]{image/graph_visualization_with_mad/20_molbace_0.51_gt.pdf}}
         & {\includegraphics[scale=0.25]{image/graph_visualization_with_mad/35_molbace_0.68_gt.pdf}} \\
         \midrule

        { \begin{turn}{90}DiGress\end{turn} }
         & {\includegraphics[scale=0.25]{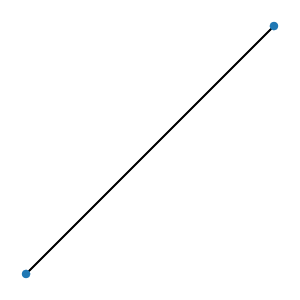}}
         & {\includegraphics[scale=0.25]{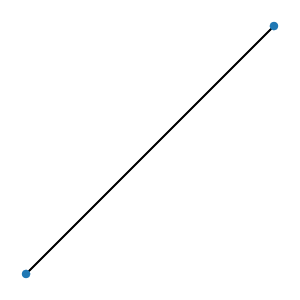}}
         & {\includegraphics[scale=0.25]{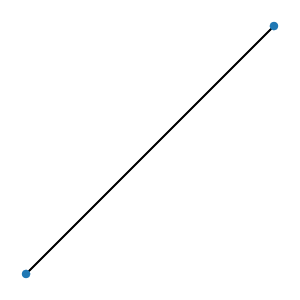}}
         & {\includegraphics[scale=0.25]{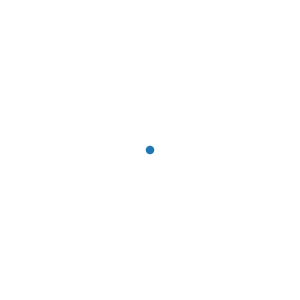}}
         & {\includegraphics[scale=0.25]{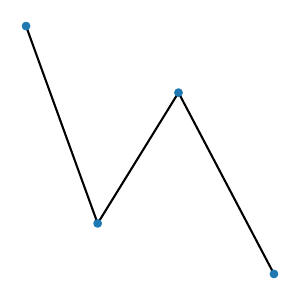}}
         & {\includegraphics[scale=0.25]{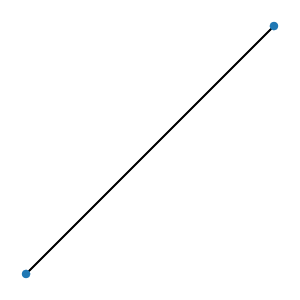}}
         & {\includegraphics[scale=0.25]{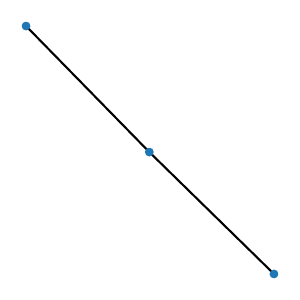}}
         & {\includegraphics[scale=0.25]{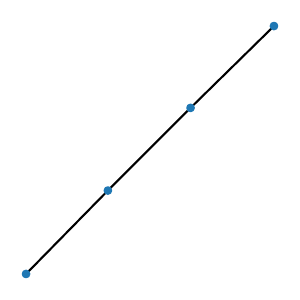}}
         &{\includegraphics[scale=0.25]{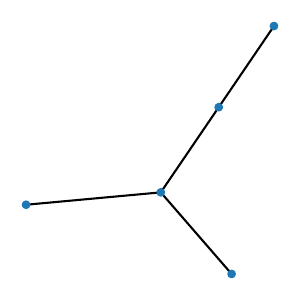}}
         & {\includegraphics[scale=0.25]{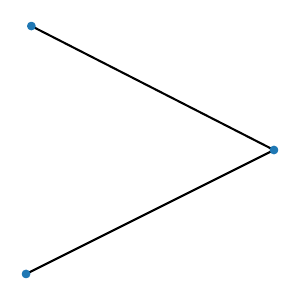}} \\

         \begin{turn}{90}\evalmetric\end{turn}
         & 0.65 & 0.50 & 0.81 & 0.61 & 0.88 & 0.86 & 0.78 & 0.77 & 0.79 & 0.96\\

        \midrule

        { \begin{turn}{90}GruM\end{turn} }
         &  {\includegraphics[scale=0.25]{image/graph_visualization_with_spectral_grum/grum_1102_wordnet_10.86_pred_spec.pdf}}
         & {\includegraphics[scale=0.25]{image/graph_visualization_with_spectral_grum/grum_2060_wordnet_5.83_pred_spec.pdf}}
         & {\includegraphics[scale=0.25]{image/graph_visualization_with_spectral_grum/grum_21_citeseer_20.24_pred_spec.pdf}}
         & {\includegraphics[scale=0.25]{image/graph_visualization_with_spectral_grum/grum_30_citeseer_17.88_pred_spec.pdf}}
         & {\includegraphics[scale=0.25]{image/graph_visualization_with_spectral_grum/grum_1350_arxiv_14.49_pred_spec.pdf}}
         & {\includegraphics[scale=0.25]{image/graph_visualization_with_spectral_grum/grum_1633_arxiv_15.56_pred_spec.pdf}}
         & {\includegraphics[scale=0.25]{image/graph_visualization_with_spectral_grum/grum_7_mutag_4.1_pred_spec.pdf}}
         & {\includegraphics[scale=0.25]{image/graph_visualization_with_spectral_grum/grum_5_mutag_8.75_pred_spec.pdf}}
         &{\includegraphics[scale=0.25]{image/graph_visualization_with_spectral_grum/grum_20_molbace_14.07_pred_spec.pdf}}
         & {\includegraphics[scale=0.25]{image/graph_visualization_with_spectral_grum/grum_35_molbace_15.81_pred_spec.pdf}} \\

         \begin{turn}{90}\evalmetric\end{turn}
         & 0.52&  0.97& 0.78& 0.69 & 0.48 & 0.21 &0.33 & 0.65 & 0.54 & 0.48\\

    \midrule
           
        { \begin{turn}{90}GenStat\end{turn} }
         & {\includegraphics[scale=0.25]{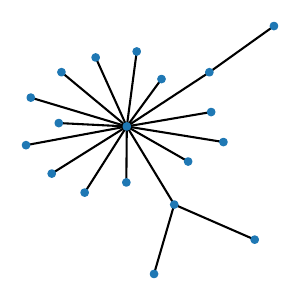}}  
         & {\includegraphics[scale=0.25]{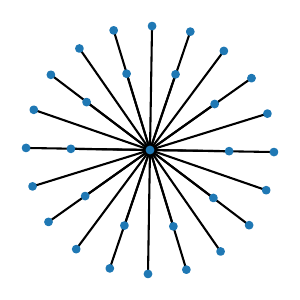}}
         & {\includegraphics[scale=0.25]{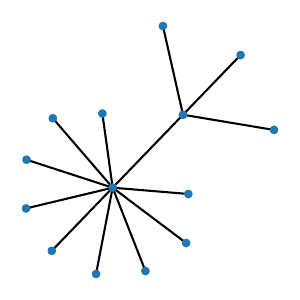}}
         & {\includegraphics[scale=0.25]{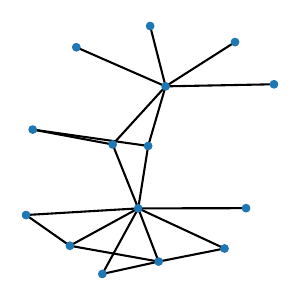}}
         & {\includegraphics[scale=0.25]{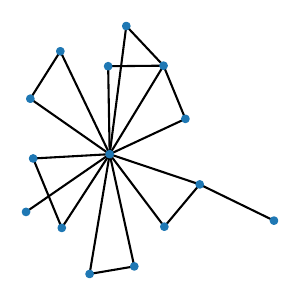}}
         & {\includegraphics[scale=0.25]{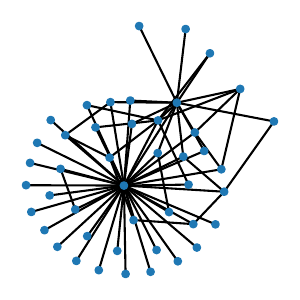}}
         & {\includegraphics[scale=0.25]{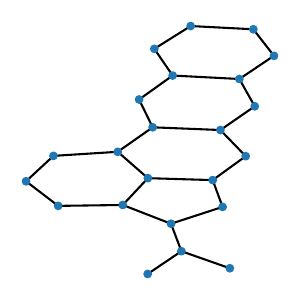}}
         & {\includegraphics[scale=0.25]{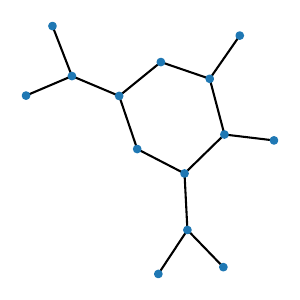}}
         &{\includegraphics[scale=0.25]{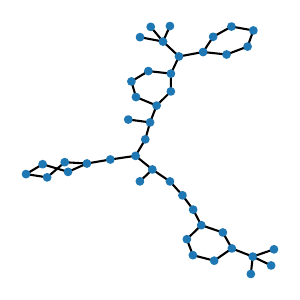}}
         & {\includegraphics[scale=0.25]{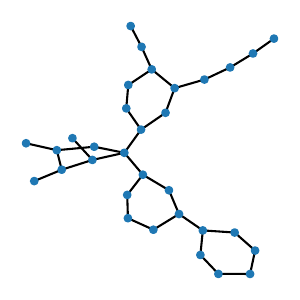}} \\

         \begin{turn}{90}\evalmetric\end{turn}
         & 0.36 & 0.48 & 0.48 & 0.49 &0.69  &0.55 &  0.36 & 0.16 & 0.19 & 0.04\\

        \midrule
        
        { \begin{turn}{90}EDGE\end{turn} }
         & {\includegraphics[scale=0.25]{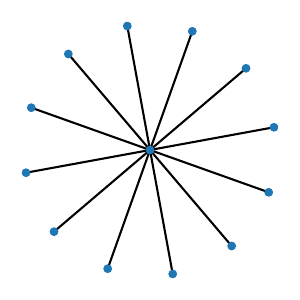}}  
         & {\includegraphics[scale=0.25]{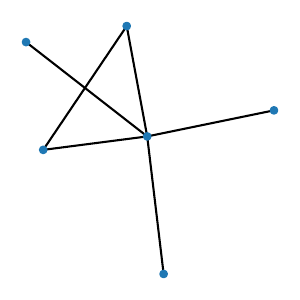}}
         & {\includegraphics[scale=0.25]{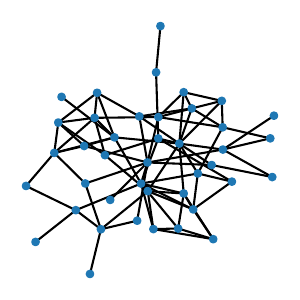}}
         & {\includegraphics[scale=0.25]{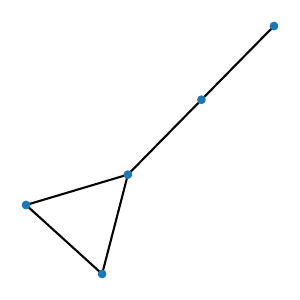}}
         & {\includegraphics[scale=0.25]{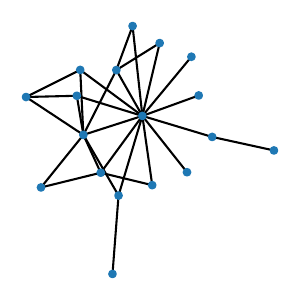}}
         & {\includegraphics[scale=0.25]{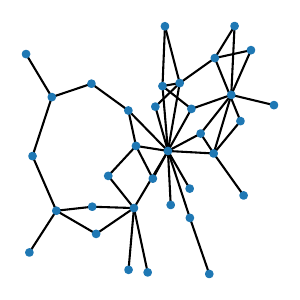}}
         & {\includegraphics[scale=0.25]{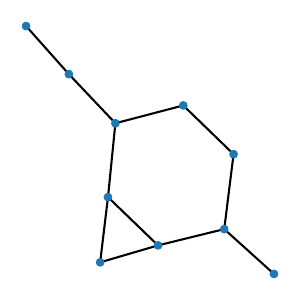}}
         & {\includegraphics[scale=0.25]{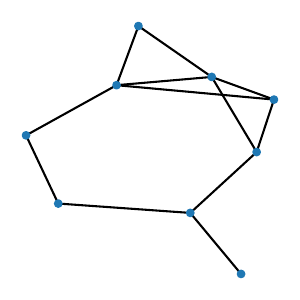}}
         &{\includegraphics[scale=0.25]{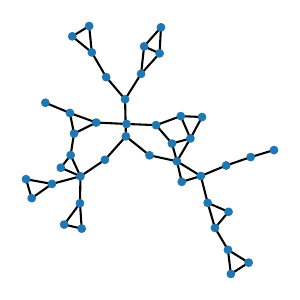}}
         & {\includegraphics[scale=0.25]{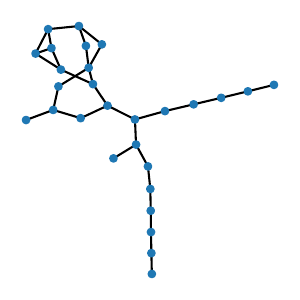}} \\

         \begin{turn}{90}\evalmetric\end{turn}
         & 0.53 & 0.20 & 0.92 & 0.83 & 0.55 & 0.14  & 0.16 & 0.53 & 0.21 & 0.10 \\
         
   \bottomrule
    \end{tabular}
    \label{tab:other_visual_result}
\end{table*}

\subsection{Attribute-Wise Error Analysis}
Table \ref{tab:attr_diff_all} shows the absolute mean error between the ground truth value and predicted value for each attribute. We use mean absolute difference (MAD$\downarrow$) metric for evaluation. MAD computes the absolute difference between the attributes of predicted graphs and their corresponding target graphs. We average these differences for each dataset. Abbreviations NC (node connectivity), EC (edge connectivity), TWMD (tree width min degree), Avg Clust (average clustering), LB (number of local bridge), Clique (number of cliques).
\label{sec:att_wise_analysis}

\begin{table*}[t]
\small
\centering
\caption{Performance of \sysname{} for each attribute across all datasets. We report the mean absolute \textbf{Difference} from target attributes; lower is better (MAD$\downarrow$ is the average over attributes).}
\begin{tabular}{lccccc}
\toprule
\textbf{Attributes} & \textbf{Citeseer} & \textbf{WordNet} & \textbf{Mutag} & \textbf{Molbace} & \textbf{Ogbn-Arxiv} \\
\midrule
\textbf{Density}       & 0.05 & 0.05 & 0.01 & 0.01 & 0.06 \\
\textbf{Edges}         & 5.44 & 4.34 & 2.77 & 4.43 & 6.85 \\
\textbf{Nodes}         & 1.89 & 3.73 & 0.00 & 2.98 & 3.68 \\
\textbf{NC}            & 0.03 & 0.00 & 0.00 & 0.00 & 0.01 \\
\textbf{Avg Clust}     & 0.16 & 0.17 & 0.03 & 0.07 & 0.19 \\
\textbf{CC}            & 0.06 & 0.06 & 0.06 & 0.04 & 0.09 \\
\textbf{LB}            & 4.16 & 7.11 & 2.33 & 6.89 & 4.14 \\
\textbf{Transitivity}  & 0.13 & 0.12 & 0.04 & 0.09 & 0.10 \\
\textbf{EC}            & 0.03 & 0.00 & 0.00 & 0.01 & 0.02 \\
\textbf{Cliques}       & 6.08 & 3.73 & 2.00 & 4.16 & 7.19 \\
\textbf{TWMD}          & 1.26 & 0.83 & 0.77 & 0.67 & 1.15 \\
\textbf{Diameter}      & 1.71 & 1.53 & 2.77 & 3.47 & 2.22 \\
\midrule
\textbf{MAD}           & \textbf{1.71} & \textbf{1.80} & \textbf{1.00} & \textbf{1.90} & \textbf{2.14} \\
\bottomrule
\end{tabular}
\label{tab:attr_diff_all}
\vspace{-8pt}
\end{table*}

\begin{table*}[t!]
\small
\centering
\caption{Performance of \sysname{} for each attribute for Ogbn-Arxiv dataset. Average mean absolute difference, MAD($\downarrow$), is the average of absolute mean error in satisfying target attributes.}
\resizebox{\textwidth}{!}{
\begin{tabular}{p{3.4cm}ccccc} 

        & \textbf{WordNet}& \textbf{Citeseer}& \textbf{Ogbn-Arxiv} & \textbf{MUTAG} &\textbf{MOLBACE}\\

        \toprule
         & \multicolumn{5}{c}{\textbf{MAD}}\\
        \midrule
       
        \textbf{GraphRNN}{\tiny~\citep{pmlr-v80-you18a}}                             & 3.26  & 5.05  &  4.80 &1.71 &3.81\\
        \midrule

        \textbf{GenStat}{\tiny~\citep{zahirnia2024neural}}  & 4.11 & 5.34 &  5.53 & 4.14 &  3.05\\
        
        \textbf{EDGE}{\tiny~\citep{chen2023efficient}}     &  3.91 & 4.97 &  5.52 &  2.62 &  3.07 \\
        
        \textbf{DiGress}{\tiny~\citep{vignacdigress}} & 5.23&6.63 & 6.67& 5.39 & 9.96 \\
        
        \textbf{GruM}{\tiny~\citep{jograph}}     & 3.75 & 5.37 & 5.42 &  3.80 &  10.8\\
        
        \midrule
        \textbf{\sysname{}}                          &\textbf{1.80} & \textbf{1.71} & \textbf{2.14} & \textbf{1.00} & \textbf{1.90}\\
        
         \bottomrule
\end{tabular}
}

\label{tab:mad_mmd_result}

\end{table*}

\subsection{Different Graph Encoder} To study the importance of using different graph encoders, we use Graph Neural Network (GNN) \cite{xu2018how} instead of CNN in \sysname{}. As shown in Table \ref{tab:gnn_encoder}, MAD error increases.

\subsection{Novelty of Generated Graphs}
To assess the novelty of the generated graphs, we quantified the extent to which the generated graphs are structurally distinct from those seen during training. Table \ref{tab:novelty} reports the fraction of  graphs generated that are not isomorphic to any of the training graphs. This shows that our model generates structurally novel graphs that differ from the training distribution.

\begin{table}[h!]
\centering
\caption{Novelty(\%) of generated graphs generated from \sysname}
\label{tab:novelty}
\begin{tabular}{l c}
\hline
\textbf{Dataset} & \textbf{Novelty (\%)} \\
\hline
Ogbn-Arxiv & 96.02 \\
Citeseer   & 100.0 \\
MOLBACE    & 100.0 \\
MUTAG      & 100.0 \\
WordNet    & 86.84 \\
\hline

\end{tabular}
\end{table}

\subsection{Out Of Distribution control attributes for Graph Generation}
 \sysname{} uses graph reconstruction during training only to learn the conditional mapping from attributes to valid graph structures. At inference time, it relies only on the desired attributes to generate graphs. However, to assess generalization, we perform experiments to quantify if the trained model can accurately generate graphs from unseen out-of-distribution attributes—those that were not derived from the dataset used during training. To generate out-of-distribution attributes, we generated 25 random graphs through Barabási–Albert \cite{barabasi1999emergence} graph generation method  and at the inference time used their attributes as the input to our model and report the generation error, SD ($\downarrow$),  in Table \ref{tab:ood}. It shows that the model maintains low generation error even on out-of-distribution attributes, demonstrating generalization beyond the training attribute distribution.

\begin{table}[h]
\centering

\caption{Generation error, SD ($\downarrow$) of \sysname{} across datasets on out of distribution attributes.}
\label{tab:ood}

\begin{tabular}{l c}
\hline
\textbf{Model Trained on} & \textbf{SD (↓)} \\
\hline
Citeseer     & 0.37 \\
MOLBACE      & 0.48 \\
MUTAG        & 0.36 \\
Ogbn-Arxiv   & 0.36 \\
WordNet      & 0.32 \\
\hline
\end{tabular}
\end{table}

\begin{table*}[t!]
\centering
\caption{Performance of \sysname{} using GNN as a graph encoder compared with CNN as an encoder. Average mean absolute difference, MAD($\downarrow$), is the average of absolute mean error in satisfying target attributes.}
\resizebox{\textwidth}{!}{
\begin{tabular}{lccccc}

        & \textbf{WordNet}& \textbf{Citeseer}& \textbf{Ogbn-Arxiv} & \textbf{MUTAG} &\textbf{MOLBACE}   \\

        \toprule
         & \multicolumn{5}{c}{\textbf{MAD}} \\
        \midrule

        \textbf{\sysname{} w GNN}    & 5.49  & 7.37 & 5.72 & 1.84 & 5.42  \\

                \textbf{\sysname{}}  &  \textbf{1.80} & \textbf{1.71} & \textbf{2.14} & \textbf{1.00} & \textbf{1.90}  \\

         \bottomrule
\end{tabular}
}
\label{tab:gnn_encoder}
\vspace{-10pt}
\end{table*}

\subsection{Order-Invariance}
To analyze and mitigate the effect of order invariance, we re-ran our experiments using a consistent node ordering through BFS (as opposed to the random ordering in the paper) to reduce the overall number of sequences to be considered. The results in Table \ref{tab:bfs_order} show, averaged across all the datasets,  that SD error slightly reduces when BFS node ordering is considered. This is because BFS preserves locality by placing structurally related nodes close to each other in the adjacency matrix. Such locality creates coherent spatial patterns that align with CNN kernels. This suggests that incorporating a more structured traversal order could improve stability by reducing sensitivity to arbitrary node permutations. In addition, our ablation study in Table \ref{tab:obj_ablation} indicates that there is negligible increase in error when using GNNs as encoder.

\begin{table}[t]
\centering
\caption{Generation error showing comparison between BFS node ordering  and random node ordering.}
\label{tab:bfs_order}

\begin{tabular}{lcc}

\hline
\textbf{Dataset} & \textbf{SD (BFS node ordering, $\downarrow$)} & \textbf{SD (random node ordering, $\downarrow$)} \\
\hline
Arxiv     & 0.26 & 0.40 \\
Citeseer  & 0.21 & 0.27 \\
Molbace   & 0.15 & 0.09 \\
Mutag     & 0.14 & 0.10 \\
Wordnet   & 0.18 & 0.44 \\
\hline
\textbf{Average} & \textbf{0.19} & \textbf{0.26} \\
\hline

\end{tabular}

\end{table}

\subsection{MMD} 
Table \ref{tab:mmd_mutag_citeseer}, \ref{tab:mmd_arxiv}, \ref{tab:mmd_molbace_wordnet} shows the MMD error across each attribute. 
Abbreviations NC (node connectivity), EC (edge connectivity), TWMD (tree width min degree), Avg Clust (average clustering), LB (number of local bridge), CC (Closeness Centrality).

\begin{table}[ht]
\centering
\caption{MMD Results on Citeseer and MUTAG (lower is better, best in \textbf{bold})}
\resizebox{\textwidth}{!}{%
\begin{tabular}{llcccccc}
\hline
Dataset &  & \sysname & GenStat & EDGE & GruM & DiGress & GraphRNN \\
\hline
\multirow{13}{*}{Citeseer}
 & Density & \textbf{0.000} & \textbf{0.000} & 0.026 & 0.076 & 0.866 & \textbf{0.000} \\
 & Edges   & \textbf{0.000} & \textbf{0.000} & 0.117 & \textbf{0.000} & 0.878 & 0.090 \\
 & Nodes   & \textbf{0.000} & \textbf{0.000} & 0.085 & \textbf{0.000} & 0.910 & 0.015 \\
 & NC      & 0.036 & \textbf{0.000} & \textbf{0.000} & \textbf{0.000} & 0.451 & 0.000 \\
 & Avg Clust      & 0.262 & \textbf{0.025} & 0.397 & 0.170 & 0.880 & 0.401 \\
 & CC      & \textbf{0.000} & \textbf{0.000} & \textbf{0.000} & 0.132 & 0.897 & 0.081 \\
 & LB      & 0.274 & \textbf{0.071} & 0.314 & 0.168 & 0.973 & 0.100 \\
 & Transitivity       & 0.139 & \textbf{0.137} & 0.233 & 0.076 & 0.845 & 0.370 \\
 & EC      & 0.036 & 0.036 & \textbf{0.000} & \textbf{0.000} & 0.451 & 0.010 \\
 & LC      & 0.074 & \textbf{0.000} & \textbf{0.000} & \textbf{0.000} & 0.790 & \textbf{0.000} \\
 & TD      & 0.160 & \textbf{0.000} & 0.165 & \textbf{0.000} & 0.850 & 0.314 \\
 & Diameter    & 0.239 & \textbf{0.000} & 0.209 & 0.202 & 1.116 & 0.146 \\
 % & Joint   & 0.258 & \textbf{0.000} & 0.274 & 0.150 & 0.858 & 0.231 \\
\hline
\multirow{13}{*}{MUTAG}
 & Density & \textbf{0.000} & \textbf{0.000} & \textbf{0.000} & 0.970 & 1.139 & 0.063 \\
 & Edges   & \textbf{0.000} & 0.021 & 0.113 & 1.029 & 1.198 & 0.387 \\
 & Nodes   & \textbf{0.000} & \textbf{0.000} & \textbf{0.000} & 0.980 & 1.181 & 0.242 \\
 & NC      & \textbf{0.000} & \textbf{0.000} & \textbf{0.000} & \textbf{0.000} & \textbf{0.000} & \textbf{0.000} \\
 & Avg Clust      & 0.594 & \textbf{0.000} & 1.158 & \textbf{0.000} & \textbf{0.000} & \textbf{0.000} \\
 & CC      & 0.307 & 0.110 & \textbf{0.000} & 0.805 & 1.169 & 0.087 \\
 & LB      & \textbf{0.000} & 0.089 & \textbf{0.000} & 0.127 & 0.771 & 0.365 \\
 & Transitivity       & 0.580 & \textbf{0.000} & 1.185 & \textbf{0.000} & \textbf{0.000} & \textbf{0.000} \\
 & EC      & \textbf{0.000} & \textbf{0.000} & \textbf{0.000} & \textbf{0.000} & \textbf{0.000} & \textbf{0.000} \\
 & LC      & \textbf{0.000} & 0.021 & 0.179 & 1.029 & 1.198 & 0.387 \\
 & TD      & 0.477 & 0.370 & 0.263 & 0.963 & 1.096 & 0.561 \\
 & Diameter     & 0.411 & \textbf{0.000} & 0.191 & 0.718 & 1.069 & \textbf{0.000} \\
 % & Joint   & 0.371 & 0.230 & 0.318 & 0.944 & 1.150 & 0.385 \\
\hline
\end{tabular}}
\label{tab:mmd_mutag_citeseer}
\end{table}

\begin{table}[ht]
\centering
\caption{MMD Results on MOLBACE and WordNet (lower is better, best in \textbf{bold})}
\resizebox{\textwidth}{!}{%
\begin{tabular}{llcccccc}
\hline
Dataset &  & \sysname & GenStat & EDGE & GruM & DiGress & GraphRNN \\
\hline
\multirow{13}{*}{MOLBACE}
 & Density & 0.149 & \textbf{0.018} & 0.314 & 0.928 & 1.161 & 0.269 \\
 & Edges   & \textbf{0.000} & \textbf{0.000} & \textbf{0.000} & 1.206 & 1.204 & 0.454 \\
 & Nodes   & \textbf{0.000} & \textbf{0.000} & 0.099 & 1.194 & 1.190 & 0.371 \\
 & NC      & \textbf{0.000} & \textbf{0.000} & \textbf{0.000} & 0.604 & \textbf{0.000} & \textbf{0.000} \\
 & Avg Clust      & 0.724 & \textbf{0.000} & 0.806 & 0.065 & 0.067 & \textbf{0.000} \\
 & CC      & 0.297 & 0.030 & 0.577 & 0.960 & 1.223 & \textbf{0.000} \\
 & LB      & 0.182 & 0.100 & 0.372 & 0.989 & 0.894 & 0.318 \\
 & Transitivity      & 0.747 & \textbf{0.000} & 0.875 & 0.064 & 0.064 & \textbf{0.000} \\
 & EC      & \textbf{0.000} & \textbf{0.000} & \textbf{0.000} & 0.604 & \textbf{0.000} & \textbf{0.000} \\
 & LC      & 0.074 & \textbf{0.000} & 0.305 & 1.164 & 1.201 & 0.450 \\
 & TWMD      & 0.592 & \textbf{0.000} & 0.714 & 1.122 & 1.099 & 0.544 \\
 & Diameter     & 0.290 & 0.036 & 0.568 & 1.192 & 1.197 & \textbf{0.000} \\
 % & Joint   & 0.684 & \textbf{0.049} & 0.852 & 0.983 & 1.177 & 0.463 \\
\hline
\multirow{13}{*}{WordNet}
 & Density & 0.194 & \textbf{0.000} & 0.144 & 0.407 & 1.119 & 0.224 \\
 & Edges   & 0.108 & \textbf{0.000} & 0.125 & 0.382 & 0.899 & 0.224 \\
 & Nodes   & 0.120 & \textbf{0.000} & 0.130 & 0.414 & 0.894 & 0.214 \\
 & NC      & 0.002 & 0.000 & 0.004 & 0.065 & 0.026 & \textbf{0.000} \\
 & Avg Clust      & 0.602 & \textbf{0.000} & 0.038 & 0.169 & 0.216 & 0.153 \\
 & CC      & 0.380 & \textbf{0.000} & 0.110 & 0.345 & 1.045 & 0.164 \\
 & LB      & 0.283 & \textbf{0.000} & 0.129 & 0.434 & 0.885 & 0.179 \\
 & Transitivity       & 0.679 & \textbf{0.000} & 0.080 & 0.203 & 0.180 & 0.128 \\
 & EC      & 0.004 & 0.000 & 0.004 & 0.065 & 0.026 & \textbf{0.000} \\
 & LC      & 0.085 & \textbf{0.000} & 0.134 & 0.403 & 0.888 & 0.209 \\
 & TWMD      & 0.613 & \textbf{0.000} & 0.008 & 0.143 & 0.324 & 0.207 \\
 & Diameter     & 0.549 & 0.014 & 0.072 & 0.098 & 0.612 & \textbf{0.017} \\
 % & Joint   & 0.580 & \textbf{0.000} & 0.139 & 0.351 & 1.012 & 0.218 \\
\hline
\end{tabular}}
\label{tab:mmd_molbace_wordnet}
\end{table}

\begin{table}[ht]
\centering
\caption{MMD Results on ArXiv (lower is better, best in \textbf{bold})}
\resizebox{\textwidth}{!}{%
\begin{tabular}{llcccccc}
\hline
Dataset &  & \sysname & GenStat & EDGE & GruM & DIiress & GraphRNN \\
\hline
\multirow{13}{*}{Ogbn-Arxiv}
 & Density & 0.069 & \textbf{0.012} & 0.092 & 0.224 & 1.004 & 0.073 \\
 & Edges   & 0.046 & \textbf{0.011} & 0.091 & 0.190 & 0.834 & 0.137 \\
 & Nodes   & 0.040 & 0.019 & 0.072 & \textbf{0.000} & 0.905 & 0.096 \\
 & NC      & \textbf{0.000} & \textbf{0.000} & 0.001 & 0.009 & 0.158 & 0.003 \\
 & Avg Clust      & 0.476 & \textbf{0.000} & 0.198 & 0.852 & 0.850 & 0.355 \\
 & CC      & 0.343 & \textbf{0.000} & 0.150 & 0.328 & 1.030 & 0.097 \\
 & LB      & 0.169 & \textbf{0.000} & 0.213 & 0.397 & 0.872 & 0.061 \\
 & Transitivity       & 0.100 & \textbf{0.000} & 0.136 & 0.772 & 0.771 & 0.287 \\
 & EC      & \textbf{0.000} & \textbf{0.000} & 0.006 & 0.011 & 0.158 & 0.005 \\
 & LC      & 0.067 & 0.018 & 0.075 & 0.066 & 0.839 & \textbf{0.077} \\
 & TWMD     & 0.125 & 0.023 & 0.071 & 0.696 & 0.747 & 0.231 \\
 & Diameter     & 0.598 & 0.014 & 0.324 & \textbf{0.017} & 0.990 & 0.205 \\
 % & Joint   & 0.433 & \textbf{0.005} & 0.252 & 0.523 & 0.984 & 0.225 \\
\hline
\end{tabular}}
\label{tab:mmd_arxiv}
\end{table}

% \begin{table}[]
%     \centering
%     \begin{tabular}{cccccccc}
%          \textbf{Dataset} &  & \sysname & GenStat & EDGE & GruM & DiGress & GraphRNN \\
%          \hline
%          \multirow{3}{*}{Citeseer} & Degree &0.001&0.001 & 0.002&0.001 &0.006 &0.001\\
%           & Clustering &0.002&0.002 & 0.003&0.002 &0.002 &0.002\\
%            & Orbit &0.026&0.020 & 0.034 & 0.013& 0.437&0.036 \\
%          \hline
%          \multirow{3}{*}{MUTAG} & Degree &0.005 &0.003 & 0.007& 0.012&0.041 &0.005\\  
%           & Clustering &0.001 &0.000 & 0.023& 0.000&0.000  & 0.000\\
%            & Orbit &0.177 &0.011 & 0.063& 0.177&0.286 &0.031 \\
%          \hline
%          \multirow{3}{*}{MOLBACE} & Degree &0.000 &0.000 &0.000 & 0.009&0.003 &0.000\\  
%           & Clustering &0.000 & 0.000& 0.002& 0.000&0.000 &0.000\\
%            & Orbit & 0.054& 0.000&0.067 & 0.322&0.287&0.033\\
%          \hline
%          \multirow{3}{*}{Ogbn-Arxiv} & Degree &0.000 & 0.000& 0.000& 0.000& 0.000& 0.000\\  
%          & Clustering &0.000 &0.000 &0.000 &0.000 &0.000 &0.000\\
%           & Orbit &0.083 &0.000 &0.014 & 0.119& 0.571&0.028 \\
%          \hline
%          \multirow{3}{*}{WordNet} & Degree &0.000 & 0.000& 0.000& 0.000&0.000 & 0.000\\  
%           & Clustering & 0.000& 0.000& 0.000& 0.000&0.000 & 0.000\\
%            & Orbit &0.057 & 0.000& 0.017&0.170 &0.572&0.035\\
%          \hline
         
%     \end{tabular}
%     \caption{Caption}
%     \label{tab:placeholder}
% \end{table}

% %%%%%%%%%%%%%%%%%%%%%%%%%%%%%%%%%%%%%%%%%%%%%%%%%%%%%%%%%%%%

\newpage

\end{document}